\documentclass[10pt,a4paper]{article}

\usepackage{geometry}
\usepackage{amsmath,amssymb}
\usepackage{booktabs}
\usepackage{array}
\usepackage{xcolor}
\usepackage[T1]{fontenc}
\usepackage[utf8]{inputenc}
\usepackage{tgtermes}
\usepackage{tgheros}
\usepackage[hidelinks,hypertexnames=false]{hyperref}
\usepackage{microtype}
\usepackage{graphicx}
\graphicspath{{figs/}}
\usepackage{tikz}
\usetikzlibrary{arrows,positioning,fit,calc}
\usepackage{needspace}
\usepackage{caption}
\usepackage{algorithm}
\usepackage{algpseudocode}

\newcommand{\logit}{\operatorname{logit}}
\newcommand{\Normal}{\mathcal N}
\newcommand{\Data}{\mathcal D}
\newcommand{\bbeta}{\boldsymbol{\beta}}

\definecolor{rule}{HTML}{9AA5AD}
\definecolor{lab}{HTML}{4A5A66}

\newcounter{sec}
\newcounter{subsec}
\renewcommand{\thesubsec}{\arabic{sec}.\arabic{subsec}}

\newcommand{\sechead}[1]{%
  \par%
  \vspace{0.9em}%
  {\sffamily\bfseries\large\color{lab}\arabic{sec}\hspace{0.65em}#1}\par
  \vspace{-0.3em}}

\newcommand{\subhead}[1]{%
  \par%
  \vspace{0.55em}%
  {\sffamily\bfseries\color{lab}\thesubsec\hspace{0.6em}#1}\par
  \vspace{-0.35em}}

\newcommand{\engsec}[1]{%
  \stepcounter{sec}\setcounter{subsec}{0}%
  \sechead{#1}}

\newcommand{\engsub}[1]{%
  \stepcounter{subsec}%
  \subhead{#1}}

\newcommand{\runin}[1]{\textbf{#1}\hspace{0.5em}}

\newcounter{app}
\newcommand{\appheadfmt}[2]{%
  \par%
  \vspace{0.9em}%
  {\sffamily\bfseries\large\color{lab}#1\hspace{0.65em}#2}\par
  \vspace{-0.3em}}
\newcommand{\engapp}[1]{%
  \stepcounter{app}%
  \appheadfmt{Supplement \Alph{app}}{#1}}

\begin{document}

\thispagestyle{empty}

\begin{figure*}[!t]
\centering
\includegraphics[width=\linewidth]{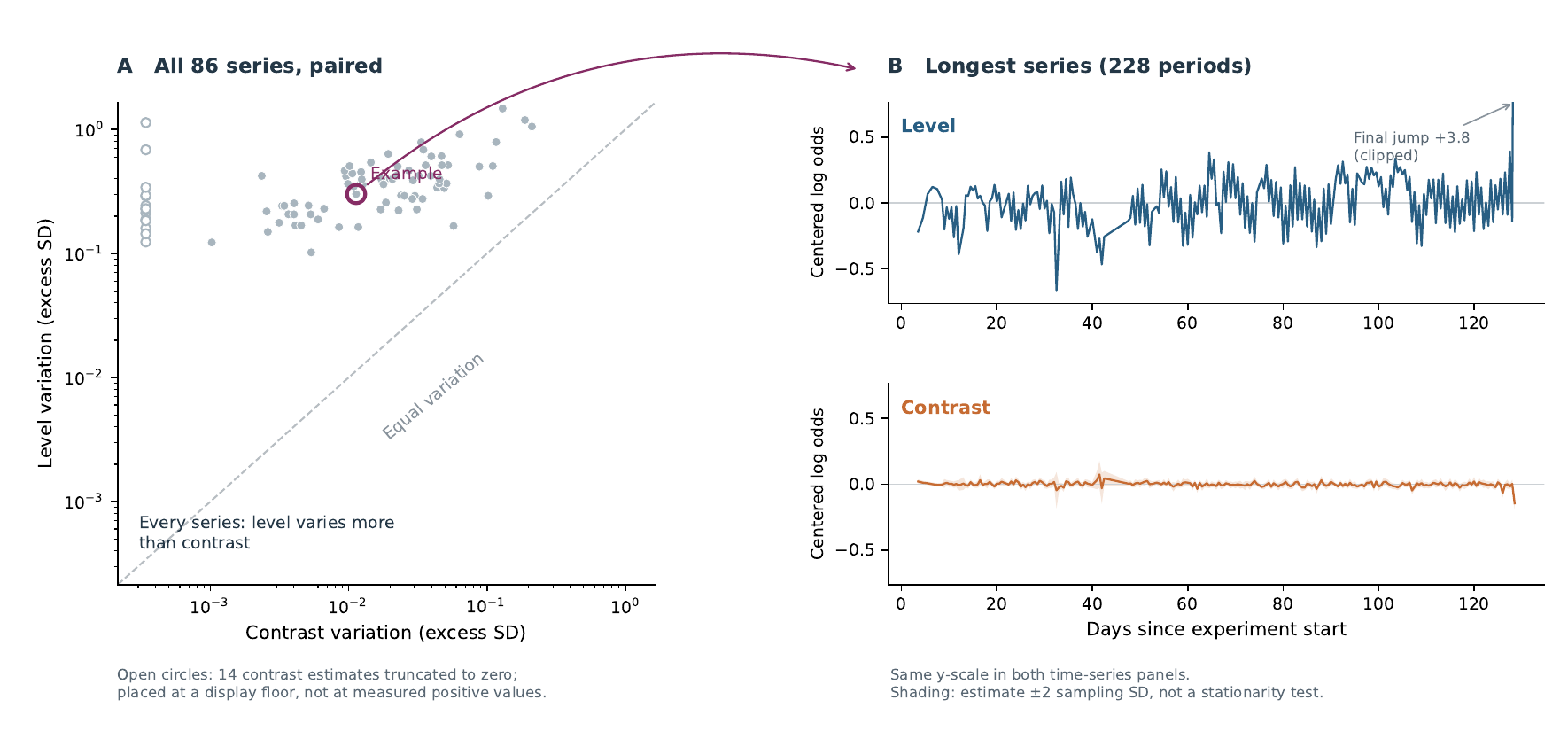}
\caption{ASOS variation, from all series to one example. A: each point pairs the excess standard deviation of the contrast (horizontal) and the level (vertical) for one of 86 series. Every series lies above the equal-variation diagonal. Open circles represent 14 contrast estimates truncated to zero, placed at a display floor rather than at measured positive values. The highlighted point and arrow identify the longest series, chosen by length. B: that series' 228 twelve-hour periods, with level above and contrast below on the same y-scale, centered at their respective means before the final period. Contrast shading is the estimate $\pm2$ sampling sd, not a stationarity test. The final level jump is clipped and annotated; all periods enter the excess-sd calculation. These diagnostics describe observed variation, not policy performance or an optimal decay rate; detection floors and provenance are in Supplement~E.}
\label{fig:asos}
\label{fig:plane}
\end{figure*}

\twocolumn[{%
\begin{center}
  {\Large\bfseries Odds-Ratio Thompson Sampling}\\[0.35em]
  {\large A Specification and Design Guide for Contrast-Based Multi-Armed Bandits}\\[0.9em]
  {\footnotesize Sulgi Kim}\\[0.15em]
  {\footnotesize\itshape Independent researcher}\\[0.15em]
  {\footnotesize\texttt{sulgik@gmail.com}}
\end{center}

\vspace{0.4em}
\textcolor{rule}{\rule{\linewidth}{0.8pt}}
\vspace{0.6em}
\begin{center}\begin{minipage}{0.94\linewidth}\small
\noindent\textbf{Abstract.}\hspace{0.5em}Batched multi-armed bandits update on a service's own schedule, and the usual implementation carries each arm's absolute reward rate from one update to the next. When the shared level moves between batches, that memory goes stale even though the comparisons between arms may not have. Odds-Ratio Thompson Sampling (OR-TS) instead carries the joint posterior over log-odds contrasts and fits the common level afresh in every batch, marginalizing it out. This paper specifies that update, places it inside a Bayesian bandit agent with two controls, decay for how much past evidence survives an update and aggressiveness for how sharply belief becomes allocation, and evaluates it against absolute-rate memory. Across 86 public A/B series the level varies about twenty-five times more than the contrast. In prespecified synthetic environments a moving level costs absolute-rate memory five times the regret and leaves the best arm below a majority of traffic in 7 of 20 runs, against none for OR-TS. In a policy simulation built from 71 real experiments, where the contrasts are too small to resolve, expected-click differences stay within $0.1\%$ for 58 of them, yet contrast memory still ends on the better arm more than twice as often. Where the contrasts themselves move, the bet fails, and that case is reported too.
\end{minipage}\end{center}
\vspace{0.6em}
\begin{center}
\includegraphics[width=0.88\linewidth]{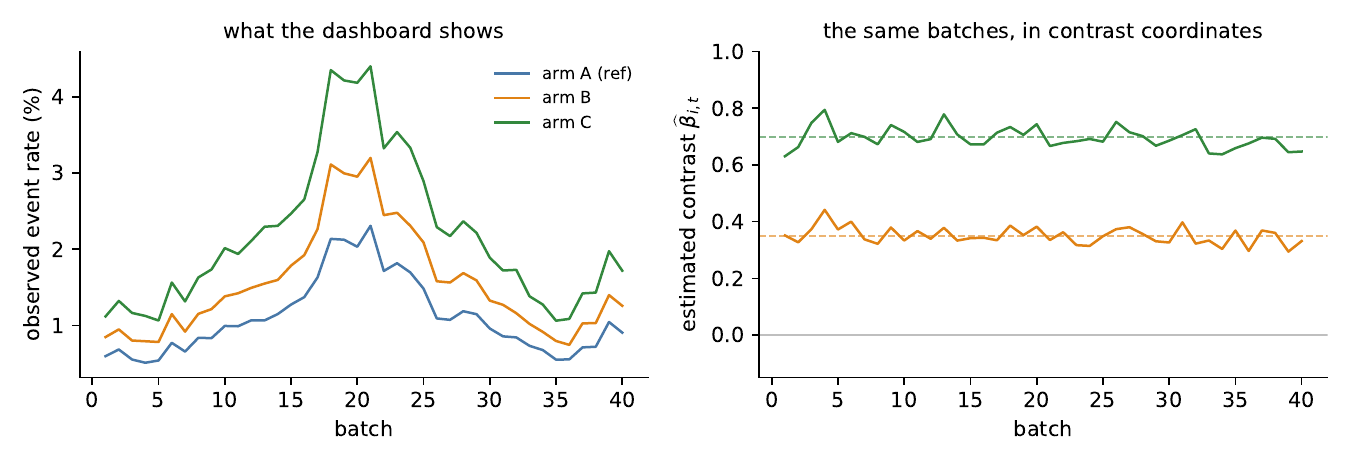}
\end{center}
\vspace{-0.4em}
\captionof{figure}{An illustrative experiment, seen twice: a constructed example with a chosen level trajectory and fixed contrasts, the Level-varying environment of Section~3; it is not fitted to a real experiment. Left: observed event rates for three arms wobble almost in parallel, because a common level dominates every curve. Right: the same batches in contrast coordinates---each $\widehat\beta_{i,t}$ is quiet around its flat true value (dashed). Stability does not mean identical estimates; it means scatter the size sampling noise predicts. Section~4.2 shows this signature on real experiments (Figure~\ref{fig:asos}).}
\label{fig:model}
\vspace{0.9em}

\vspace{0.9em}
}]

\runin{Keywords.}
Multi-armed bandit; Thompson sampling; A/B testing; adaptive experiment; selective posterior persistence; treatment contrast; odds ratio; logistic regression; temporal drift; posterior decay.

\engsec{Introduction}

\runin{Sequential decisions in online services.} Online services repeatedly choose which option to show and how much traffic to give it. A/B tests compare these options under a planned allocation (Kohavi et al.\ 2020); a multi-armed bandit (MAB) adapts their traffic shares to accumulating evidence, an \emph{adaptive allocation} (Lai and Robbins 1985; Bubeck and Cesa-Bianchi 2012). Textbook bandits revise after every observation, but a service revises on its own schedule, so this paper treats the \emph{batched} bandit: one update serves a whole \emph{batch} (Section~2) of traffic, and the batch is the service's operational update cycle, often daily or weekly (LaunchDarkly 2026b). It is also the unit of comparison: its arms face the same period's conditions, so what a later batch can inherit is the comparison, not the rates. The step from comparison to adaptive allocation is most direct when rewards are observed promptly and earlier exposures have negligible carryover into later batches. In that setting, a service can observe responses to the competing options, called arms, revise its belief, and allocate the next batch.

The allocation can remain balanced, as in an A/B test, or favor promising arms as evidence accumulates. For adaptive allocation, the objective here is to minimize \emph{cumulative regret}, the expected reward forgone relative to always selecting the best arm (Scott 2010; Russo et al.\ 2018). We use Thompson sampling, which gives each arm a share equal to its current probability of being best (Thompson 1933). Even without carryover from earlier exposures, the environment can change between updates, making some of the evidence behind those probabilities outdated. \emph{What, then, should be carried into the next update?} This is the question of this paper; we call the belief a bandit carries forward its \emph{memory}.

\runin{One experiment, two descriptions.} Consider a two-arm bandit at one weekday update, with arm 1 at an event rate of 10\% and arm 2 at 12\%. The usual bandit stores two absolute facts: arm 1 is at 10\% and arm 2 is at 12\%. The same observations can be written instead as one \emph{current level}, arm 1's rate, and one \emph{relative effect}, arm 2's \emph{odds ratio} (Section~2.1) against arm 1, about $1.23$, close to the $20\%$ lift a practitioner would quote at rates this small. Both descriptions hold the same rates and the same ranking; they differ in what their coordinates mean, and the weekend shows why that matters. Let the event be a purchase, and suppose the weekend brings more visitors who are browsing rather than ready to buy, lowering arm 1's rate to 5\% while the odds ratio stays where it was; arm 2 is then at about 6.1\%. The stored rates of 10\% and 12\% no longer describe the current level, but the comparison still favors arm 2, and the comparison is all the next allocation needs. Figure~\ref{fig:model} draws the same distinction in a separate three-arm example.

\runin{From comparisons to allocation.} A treatment contrast parameterizes a comparison between arms. Here the contrast is a difference in log odds, so its exponential is an odds ratio. This specifies the comparison being learned, the \emph{estimand}, the quantity the analysis sets out to learn; it is not numerically interchangeable with a rate difference or the relative lift a platform reports. Reference-coded logistic regression separates the log odds of one arm, the \emph{reference arm}, which any arm can be, from the contrasts of the others against it (Agresti 2003). The contrasts determine the ranking used for allocation; translating them into an absolute reward difference also requires the level.

Yet standard bandit practice carries beliefs about each arm's absolute reward rate across cycles. For binary rewards the common implementation is an independent Beta posterior per arm, which we call Beta--Bernoulli Thompson sampling, \emph{Beta-TS} (Russo et al.\ 2018): two counters per arm, easy to deploy, and the default in representative tutorials (Scott 2010; Chapelle and Li 2011; Russo et al.\ 2018) and in experimentation platforms (Optimizely 2026a; LaunchDarkly 2026a; McKenzie 2020), which themselves warn that time-correlated conversion rates can skew the allocation (Adobe 2026). Beta-TS thereby also carries the background level the arms share, folded into every arm's posterior and held fixed for the whole run, although a joint belief about the relative effects and their uncertainty is all that each arm's probability of being best requires (Section~2.2).

\runin{Remember comparisons; relearn the baseline.} Fit the common level anew within each batch, integrate it out as a \emph{nuisance parameter} (Section~2.2), and carry the joint posterior over arm comparisons to the next update and allocation. This is the contrast-based memory rule. With binary rewards and a logistic link the comparisons are log-odds contrasts, and exponentiating them gives odds ratios; hence the name \emph{Odds-Ratio Thompson Sampling (OR-TS)}, retained from Kim and Kim (2020). In the terms of the loop, which this paper calls \emph{recognition} and \emph{action}---recognition updates the posterior over the reward parameters, action allocates on it (\AA str\"om 1965; Kaelbling et al.\ 1998)---the part of the posterior carried from one cycle to the next is the algorithm's \emph{internal state}, the memory asked about above. Its unit is the batch: the level the arms share within a batch is what lets the level be separated from the arm effects (Section~2).

\begin{figure*}[t]
\centering
\centering
\resizebox{0.62\textwidth}{!}{
\begin{tikzpicture}[>=stealth,thick,
 box/.style={draw,rounded corners=2pt,minimum width=40mm,minimum height=21mm,align=center,font=\normalsize}]
\node[box,fill=black!18] (state) at (0,0) {Recognition\\[2pt]{\small posterior belief $q$}};
\node[box,fill=black!10] (action) at (60mm,0) {Action\\[2pt]{\small allocate $\pi\propto\rho^{\gamma}$}};
\draw[->] (state.east)--(action.west);
\node[font=\small] at (30mm,4mm) {belief $q$};
\node[font=\small] at (60mm,-14mm) {aggressiveness $\gamma$};
\node[circle,draw,fill=white,inner sep=1pt,font=\normalsize] (sum) at (0,20mm) {$+$};
\draw[->] ([xshift=-7mm]state.north) -- (-7mm,15mm) .. controls (-7mm,20mm) and (-5mm,20mm) .. (sum.west);
\draw[->] (sum.east) .. controls (7mm,20mm) and (7mm,18mm) .. (7mm,15mm) -- ([xshift=7mm]state.north);
\node[font=\small,anchor=east] at (-9mm,16mm) {decay $\lambda$};
\node[font=\small,anchor=west] at (9mm,16mm) {update};
\draw[->,blue!55!black] (action.north) -- (60mm,26mm) -- (0,26mm) -- (sum.north);
\node[font=\small,text=blue!55!black,fill=white] at (30mm,26mm) {data $\mathcal D_t$};
\end{tikzpicture}}
\caption{A Bayesian bandit agent with two controls. The upper loop is the update: decay $\lambda$ discounts the carried belief before the batch's information is added, a posterior update rather than an addition of cumulative counts. Recognition passes the belief to action, which draws from it and turns the resulting winner probabilities into the allocation, as sharply as aggressiveness $\gamma$ says. That allocation is served as the next batch, and the batch's data $\mathcal D_t$ return to the update: with binary rewards, the exposure and event counts per arm; under another likelihood, whatever that likelihood consumes (Section~6.2). What the belief holds is the policy's choice, and Section~5 takes it up.}
\label{fig:agent}
\end{figure*}
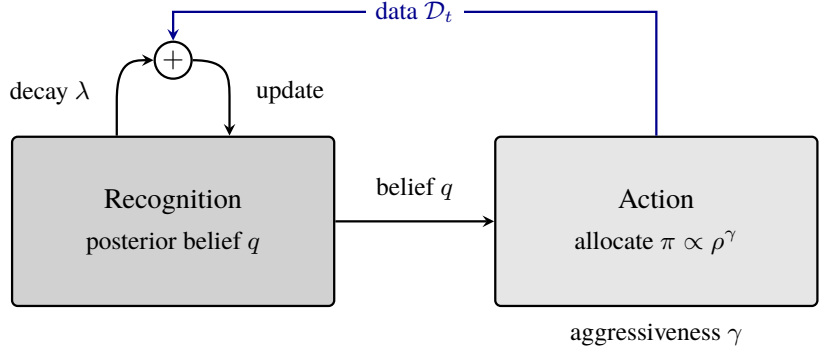

This paper specifies OR-TS for binary rewards, evaluates it, and examines its state-separation assumption using data from real-world experiments. In 86 variant-versus-control series from 71 real-world online controlled experiments, the common level moves more than the contrast in every series; in registered common-shock simulations, Beta-TS ends 7 of 20 runs with the best arm below a majority of traffic, and OR-TS ends none that way. The memory rule is further extended with one explicit control for each of the two questions that follow it within the loop, \emph{decay} $\lambda$ for how much of the memory to retain and \emph{aggressiveness} $\gamma$ for how strongly the current belief should guide action (Section~5); both arose from practical needs, and Figure~\ref{fig:agent} draws the resulting loop, which any Bayesian bandit could run.

Written as a specification and design guide for practitioners and platform engineers, the paper emphasizes the meaning of the state, the operational update, and its assumptions and diagnostics, and defers derivations to the supplementary material. The comparisons distinguish Stationary environments, Level-varying (level changes with fixed contrasts), and Contrast-varying (changing contrasts); Section~3 defines these cases.

\runin{Where this fits in a platform.} OR-TS specifies the traffic-allocation update for a batched binary-reward bandit. It assumes that the recorded trials support the binomial likelihood, arms share a credible level within each batch, and historical contrasts remain useful over the chosen memory horizon. The studied setting has prompt outcomes and negligible carryover between exposures. Table~\ref{tab:migration} shows the integration boundary; Supplement~D says what the platform must supply and how the state is recovered.

\par\medskip
\noindent\begin{minipage}{\columnwidth}
\centering\small
\begin{tabular}{@{}p{0.19\linewidth}p{0.33\linewidth}p{0.37\linewidth}@{}}
\toprule
\textbf{Component} & \textbf{Beta-TS} & \textbf{OR-TS} \\
\midrule
Update & Two counters per arm & Logistic fit, marginalize, $M$ draws \\
\addlinespace[0.3em]
State & $2K$ counters & $K{-}1$ contrast means and a $(K{-}1)\times(K{-}1)$ precision matrix \\
\bottomrule
\end{tabular}
\captionof{table}{What migration changes. The interface is untouched: the same per-batch counts go in and the same winner probabilities come out. Inside, the state grows from $2K$ counters to a contrast posterior of about $K^2/2$ numbers, and that state is rebuilt from the per-batch log, not from running totals. Supplement~D gives the compute cost and the recovery rules.}
\label{tab:migration}
\end{minipage}\par\medskip

\noindent\emph{Reading paths.} To implement the method, Section~2 (Algorithm~1 and Table~\ref{tab:migration}), Supplement~C (the full algorithm with recovery) and Supplement~D (migration) suffice; which method suits which situation is Table~\ref{tab:situations} in Section~7. The evidence is in Section~4 and Supplement~F; the derivations in Supplements~A and~B.\par\medskip

\engsec{The procedure}

Section~2.1 follows one update from incoming counts to the next traffic allocation. Algorithm~1 collects the steps; Section~2.2 explains why the decision can use the marginal contrast posterior. Supplement~A gives the calculations and initialization conditions.

\runin{Problem setup: a batched binary-reward bandit.} The box fixes the notation and the one cycle it describes.

\vspace{-0.35em}
\noindent\begin{minipage}{\linewidth}
\textcolor{rule}{\rule{\linewidth}{0.5pt}}\\[0.1em]
{\textbf{The setting.} One cycle of a batched binary-reward bandit}\\[0.1em]
\textcolor{rule}{\rule{\linewidth}{0.4pt}}
\small
\begin{tabular}{@{}p{0.19\linewidth}p{0.73\linewidth}@{}}
$K$ & arms, active in batch $t$ \\[0.1em]
batch $t$ & one update cycle, and the unit of comparison: its arms share the period's conditions \\[0.1em]
$\pi_{i,t}$ & the share of batch-$t$ traffic sent to arm $i$, held fixed until the batch closes; $\sum_i\pi_{i,t}=1$ \\[0.1em]
$p_{i,t}$ & arm $i$'s unknown event probability in batch $t$ \\[0.1em]
$n_{i,t},c_{i,t}$ & arm $i$'s exposures and events, recorded when the batch closes \\[0.1em]
boundary $t$ & the instant batch $t$ closes, the policy updates, and batch $t+1$ opens; every step of OR-TS runs there (Figure~\ref{fig:state}) \\[0.1em]
\end{tabular}
\[
\begin{aligned}
c_{i,t}\mid n_{i,t},p_{i,t}&\sim\operatorname{Binomial}(n_{i,t},p_{i,t}),\\[0.15em]
\{(n_{i,t},c_{i,t})\}_{i=1}^K \;\longrightarrow&\; \mathcal B_t \;\longrightarrow\; \pi_{t+1}.
\end{aligned}
\]
Here $\mathcal B_t$ is the belief state the policy carries across the boundary. Conditioning on the realized exposures gives that likelihood whether the platform sets integer quotas $n_{i,t}\approx N_t\pi_{i,t}$ for a known batch size or routes each arriving unit to arm $i$ with probability $\pi_{i,t}$, so nothing below depends on which mode is used. Two counts per arm are all the likelihood needs, and outcomes observed during batch $t{+}1$ affect routing only at the next boundary.
\noindent
\textcolor{rule}{\rule{\linewidth}{0.5pt}}
\end{minipage}
\vspace{-0.2em}

\engsub{From reference coding to OR-TS}

At boundary $t$ the platform hands over two counts per arm, exposures $n_{i,t}$ and events $c_{i,t}$, for the batch that has just closed. OR-TS fits one logistic regression to those counts together with the contrast posterior carried from the previous boundary, keeps the contrast part of the fitted posterior as the new state, and draws from that state to set the next batch's traffic shares. Each batch's counts enter once: earlier observations are already summarized in the carried posterior, so their counts are not added again (Supplement~D). The rest of this subsection names the parts.

Index the arms as $1,2,\ldots,K$ and write their batch-$t$ event probabilities in the reference-coded logistic form (Agresti 2003)
\begin{equation}
\logit(p_{i,t})
=\alpha_t+\sum_{j=1}^{K-1}\beta_j\mathbf 1\{i=j\},
\qquad i=1,\ldots,K.
\label{eq:reference_model}
\end{equation}
This is an ordinary logistic regression on dummy-coded arm indicators. Arm $K$ carries the all-zero indicator row and is the \emph{reference arm}. The name means only that arm $K$ is the origin from which the other arms' contrasts are measured: it need not be the control, the incumbent, or the current leader, any arm can serve, and the allocation does not depend on which one is chosen (Supplement~B). Section~1's example used arm 1 as the origin. The intercept $\alpha_t$ is the current log odds of arm $K$ and is the level common to all arms in batch $t$; each coefficient $\beta_j$ is arm $j$'s log-odds contrast against that reference, and $\bbeta=(\beta_1,\ldots,\beta_{K-1})$ is the joint contrast vector. Only the intercept carries a batch index: a fresh $\alpha_t$ is fitted at every boundary, whereas $\bbeta$ is the single vector the state carries from batch to batch. This builds Section~3's Level-varying model into the update: $\alpha_t$ free, $\bbeta$ carried. That state is one object, not $K{-}1$ separate odds ratios: the joint posterior of all the contrasts with their covariance. Every contrast is measured against the same reference arm, so the contrasts share that arm's noise and are correlated; step A1 therefore draws them together, and a per-arm marginal would misstate the winner probabilities. Supplement~B shows that this one object can be re-expressed against any reference or restricted to any subset of arms by a linear map.

\runin{Recognition: carry the contrasts, not the level.}
The newly observed outcomes in batch $t$ are aggregated into the exposure and event counts $\{(n_{i,t},c_{i,t})\}_{i=1}^K$. At boundary $t$, recognition combines these new data with the joint contrast posterior $q_{t-1}(\bbeta)$ carried from the preceding boundary, represented in the basic implementation by a Gaussian with mean $\mu_{t-1}$ and precision $S_{t-1}$ (the inverse covariance). It gives $\alpha_t$ a fresh flat prior and forms a joint posterior for $(\alpha_t,\bbeta)$. The basic implementation uses the standard Laplace approximation for Bayesian logistic regression, a Gaussian fitted at the posterior's peak (Bishop 2006, Section~4.5.1). It then integrates out $\alpha_t$---averages over its uncertainty rather than fixing it---and retains the marginal contrast posterior
\[
q_t(\bbeta)\approx\Normal(\mu_t,S_t^{-1}),
\]
and carries only $(\mu_t,S_t)$ to the next boundary.

\runin{Action: sample the contrast posterior, allocate by its winners.}
Where independent Beta-TS samples each arm's success rate separately, OR-TS \emph{jointly samples the entire contrast vector} from $q_t(\bbeta)$, preserving correlations between contrasts. Sampling their marginals independently would discard dependence induced by the shared reference and change the winner probabilities. Choosing the winner of each draw remains ordinary Thompson sampling (Thompson 1933; Scott 2010; Russo et al.\ 2018). Append a score of $0$ for reference arm $K$; the fraction of draws won by arm $i$ gives its next-batch allocation probability $\pi_{i,t+1}$. For example, with arm 3 as reference, $(\beta_1,\beta_2)=(0.30,-0.10)$ gives scores $(0.30,-0.10,0)$, so arm 1 wins; arm 3 wins when both contrasts are negative. If the arms win 62\%, 8\%, and 30\% of $M$ draws, those are the next batch's shares. Action needs nothing but draws from this joint posterior, no separate point estimate or exploration bonus. The arm set is fixed here; Section~6.1 relaxes this.

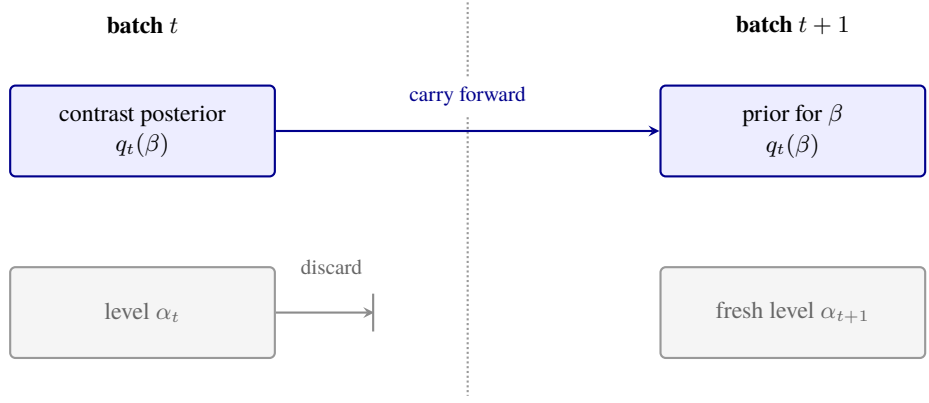
\begin{figure*}[!t]
\centering
\begin{tikzpicture}[>=stealth,thick,
 box/.style={draw,rounded corners=2pt,minimum width=35mm,minimum height=12mm,align=center,font=\small},
 level/.style={box,draw=black!40,text=black!60,fill=black!4},
 state/.style={box,draw=blue!55!black,fill=blue!7}]
\node[font=\small\bfseries] at (0,14mm) {batch $t$};
\node[font=\small\bfseries] at (86mm,14mm) {batch $t+1$};
\node[level] (a) at (0,-24mm) {level $\alpha_t$};
\node[level] (b) at (86mm,-24mm) {fresh level $\alpha_{t+1}$};
\node[state] (q) at (0,0) {contrast posterior\\$q_t(\beta)$};
\node[state] (prior) at (86mm,0) {prior for $\beta$\\$q_t(\beta)$};
\draw[densely dotted,black!40] (43mm,17mm)--(43mm,-35mm);
\draw[->,black!45] (a.east)--++(13mm,0);
\draw[black!45] (30.5mm,-26.5mm)--(30.5mm,-21.5mm);
\node[font=\footnotesize,text=black!60] at (25mm,-18mm) {discard};
\draw[->,blue!55!black] (q.east)--(prior.west) node[midway,above=2mm,fill=white,font=\footnotesize] {carry forward};
\end{tikzpicture}
\caption{What crosses a batch boundary in OR-TS. The fitted level $\alpha_t$ is marginalized out; only the contrast posterior $q_t(\bbeta)$ is carried into the next batch, where a fresh level $\alpha_{t+1}$ is fitted.}
\label{fig:state}
\end{figure*}

We call this boundary-level recognition--action rule \emph{Odds-Ratio Thompson Sampling (OR-TS)}, and use the same name for the sequential policy obtained by repeating it over a run. Figure~\ref{fig:state} isolates the carry-over rule: the fitted level ends at the boundary, while the contrast posterior becomes the next batch's prior.

\begin{figure*}[!t]
\centering
\begin{minipage}{0.94\linewidth}
\textcolor{rule}{\rule{\linewidth}{0.6pt}}\\[0.15em]
{\textbf{Algorithm 1.} One OR-TS boundary update, iterated over the experimental run}\\[0.1em]
\textcolor{rule}{\rule{\linewidth}{0.4pt}}
\small
\smallskip
\textbf{Before the first batch.} Choose the arm set, reference arm $K$, the effect scale $\tau$ (default $\sqrt2$, so that every pairwise log-odds contrast has prior SD $2$), and Monte Carlo budget $M$. Store the symmetric proper contrast prior $\mu_0=0$, $\Sigma_0=\tau^2(I+\mathbf1\mathbf1^\top)$ (Supplement~A). The level gets no stored prior: each batch's intercept $\alpha_t$ is fitted afresh under a flat prior (R1). Allocate the first batch uniformly or by a prespecified start-up rule.\par\smallskip
\textbf{At boundary $t$.} The platform supplies the batch's exposure and event counts $\{(n_{i,t},c_{i,t})\}$ for every active arm; the update consumes them as given.\par
\medskip

\begin{minipage}[t]{0.47\linewidth}
\textbf{1. Recognition---update the state}\par\smallskip
\hangindent=1.7em\hangafter=1
\textbf{R1}\hspace{0.6em}\emph{Fit one logistic regression.}$^{\dagger}$ Fit Equation~\eqref{eq:reference_model} with a fresh flat-prior intercept $\alpha_t$ and the carried contrast prior $(\mu_{t-1},S_{t-1})$, and form the joint Laplace approximation.\par\smallskip
\hangindent=1.7em\hangafter=1
\textbf{R2}\hspace{0.6em}\emph{Marginalize the nuisance level.} Retain the marginal contrast Gaussian $q_t(\bbeta)\approx\Normal(\mu_t,S_t^{-1})$. Save $(\mu_t,S_t)$ and discard the intercept.\par
\end{minipage}
\hfill
\begin{minipage}[t]{0.47\linewidth}
\textbf{2. Action---allocate the next batch}\par\smallskip
\hangindent=1.7em\hangafter=1
\textbf{A1}\hspace{0.6em}\emph{Query the state.} Draw $M$ contrast vectors from $\Normal(\mu_t,S_t^{-1})$; in each draw, give arm $K$ the score $0$ and identify the best arm.\par\smallskip
\hangindent=1.7em\hangafter=1
\textbf{A2}\hspace{0.6em}\emph{Allocate the next batch.} Let $\rho_{i,t}$ be the fraction of draws won by arm $i$, and use $\pi_{i,t+1}=\rho_{i,t}$ for the next batch.\par
\end{minipage}

\medskip
\textbf{Repeat the recognition--action cycle.}\hspace{0.6em}\emph{Memory rule: carry $(\mu_t,S_t)$ forward; do not carry $\alpha_t$.} Decay and aggressiveness are optional controls (Section~5; Algorithm~2, Supplement~C).\par
\smallskip
\noindent\textcolor{rule}{\rule{0.3\linewidth}{0.3pt}}\par
{\footnotesize\noindent\hangindent=1em\hangafter=1 $^{\dagger}$\,\emph{Sparse batches.} The fit is defined whenever the batch as a whole has at least one event and one non-event; an individual arm may show only one kind. If the batch lacks one, skip R1--R2 and keep the current state, or the start-up allocation if no fit has completed yet; do not pool its counts with a later batch under one intercept. The historical flat initialization ($S_0=0$) remains an explicit option and additionally requires every arm to show both kinds in the first fitted batch (Supplement~A).\par}
\noindent
\textcolor{rule}{\rule{\linewidth}{0.6pt}}
\end{minipage}
\end{figure*}

\runin{Approximation check.} OR-TS stores the contrast posterior as a Gaussian, which is exact only when events are plentiful. Supplement~G measures what the approximation does to the winner probabilities against an exact calculation for two arms (Table~\ref{tab:fidelity}): with hundreds of events per batch the error is a few hundredths of a percentage point, with tens of events up to about two points, and with a handful of events per batch several points. In very sparse batches, then, the approximation rather than the memory rule is what to watch, and Section~7 uses the table as a batch-size guide. The check does not cover adaptive allocation or larger arm sets.

\engsub{Why the contrasts are enough}

Step R2---discarding the fitted intercept while retaining the contrast posterior---is the distinctive memory rule of OR-TS. Two plain facts make it coherent.

\runin{The coordinates can represent every batch.}
Describing $K$ arms by one level and $K{-}1$ contrasts relabels their $K$ rates; it does not simplify them. In the example of Section~1, ``arm 1 at 10\%, arm 2 at 12\%'' and ``level 10\%, odds ratio 1.23'' are the same two numbers written twice, and either pair can be recovered from the other; the same holds for any $K$, because $\alpha_t$ is the reference arm's log odds and each $\beta_j$ is another arm's log odds minus it (Figure~\ref{fig:model} shows the two descriptions side by side). Equation~\eqref{eq:reference_model} can therefore represent any vector of $K$ arm rates: the summation there is dummy-coding notation, not an assumption that arms add up in any way, and the reference arm is a choice of origin and nothing more (Supplement~B gives the map between choices; Supplement~A states the fact formally). This is a statement about the parameterization, not equivalence of priors, fitted estimates, or statistical efficiency. The persistence assumption concerns how the coordinates are used across batches.

\runin{The contrasts alone pick the winner.}
Thompson sampling allocates traffic by each arm's posterior probability of being the best. Changing the level $\alpha_t$ shifts every arm's log odds by the same amount, which moves all the rates together but never reorders them, because a higher log odds is always a higher rate: in the example, when the level falls and 12\% against 10\% becomes 6.1\% against 5\%, arm 2 is still ahead. Hence
\[
\arg\max_{i=1,\dots,K} p_{i,t}
=\arg\max\{\beta_1,\ldots,\beta_{K-1},0\}
\]
at every parameter value, and so at every posterior draw: each arm's probability of being best is the same number whether it is computed with the level or without it. Step~A1 is therefore ordinary probability matching over the arm rates, carried out in contrast coordinates and jointly: the $K{-}1$ contrasts share the reference arm's noise, so their posterior correlation is part of what each draw carries. What distinguishes OR-TS is the state supplied to it. The intercept is still needed to fit the counts and to size the contrast uncertainty, but for choosing among arms it is a nuisance parameter. That is a statement about the decision, not a claim that $\alpha_t$ carries no information; Supplement~A gives the caveat.

\runin{Carry only what the decision needs.}
The marginal contrast posterior is sufficient to compute the current winner probabilities. Carrying it into the next update additionally assumes that the contrasts persist while each batch receives a fresh level; Supplement~A states that model and its exact marginal recursion. Section~3 compares its assumptions with Beta-TS, and Section~4.2 examines contrast variability in real-world experiments.

\engsec{What to expect when the environment changes}

The practical question is which part of yesterday's evidence remains useful for tomorrow's allocation. We compare OR-TS (Kim and Kim 2020), which remembers contrasts and relearns the baseline, with Beta-TS (Russo et al.\ 2018), which remembers each arm's absolute reward rate. Their stationarity assumptions are nested: fixed absolute arm rates imply fixed log-odds contrasts, whereas fixed contrasts permit a changing common level. It separates two questions: the validity of carrying absolute-rate beliefs when the level changes, and the possible efficiency cost of relearning the level when it is stable. Three named environments organize the rest of the paper. In the notation of Equation~\eqref{eq:reference_model}, let both parameters carry a batch index, $\logit(p_{i,t})=\alpha_t+\beta_{i,t}$, with $\bbeta_t$ the contrast vector of batch $t$. \emph{Stationary}: the common level and the contrasts are both fixed, $\alpha_t=\alpha$ and $\bbeta_t=\bbeta$. \emph{Level-varying}: the common level changes while the contrasts stay fixed, $\alpha_t$ free and $\bbeta_t=\bbeta$. \emph{Contrast-varying}: the contrasts change, $\bbeta_t$ free, whether or not the level does. The names identify what varies, not an ordering of drift magnitude or difficulty; the paragraphs below say what each policy should do in each, and Section~4 tests it.

\runin{Stationary.} Both policies' stationarity assumptions hold. OR-TS nevertheless fits a separate level each batch instead of pooling evidence about a single stable level, so it may lose statistical efficiency and learn the best arm more slowly. This is a possible cost, not a consequence established by nesting alone: priors and posterior approximations also differ. Section~4.1's undisturbed simulation tests whether that cost appears in allocation and cumulative regret, and separately measures the cost of unnecessary decay.

\runin{Level-varying.} In Section~1's example, both conversion rates fall while their odds ratio stays fixed. Beta-TS's fixed-rate model is misspecified: its historical absolute-rate posterior need not be valid for the current rates or their comparison under adaptive allocation. OR-TS's fixed-contrast assumption can still hold. OR-TS fits the new baseline while retaining the comparison, so historical contrast evidence remains useful without carrying an obsolete level. This is its intended advantage, not invariance of regret: event rates still affect contrast information and reward gaps (Supplement~A). Section~4.1 tests its consequences under a common shock; Section~4.2 examines the corresponding structure in real-world experiments.

Why the level hurts Beta-TS is a matter of allocation. How much baseline movement Beta-TS can absorb depends on the allocation as much as on the movement: each Beta-TS arm posterior is a cumulative average, so a common shift distorts the apparent contrast only in proportion to the difference in how much of each arm's history the latest batch contributes. Under balanced allocation that difference is nil; under the imbalance a bandit creates it grows, and the distortion with it. Section~4.1 measures how large a common shock Beta-TS absorbs before that mechanism costs it the best arm.

\runin{Contrast-varying.} Both fixed-rate and fixed-contrast assumptions can now fail. OR-TS also faces model misspecification, not merely an efficiency cost: past comparisons can become stale, even without a change of best arm. Relearning the baseline alone cannot fix this, so OR-TS with full memory can also adapt slowly. \emph{Posterior decay} weakens historical evidence before each new batch, following the discounting approach used in nonstationary UCB and Thompson sampling (Garivier and Moulines 2011; Raj and Kalyani 2017), here applied to contrasts. Its control $\lambda=0$ retains all evidence; positive values progressively forget it, with effective memory roughly $1/\lambda$ batches. Faster adaptation comes at the cost of discarding useful evidence when effects are stable. Section~5.1 gives the equations and tuning guidance; Section~6.2 discusses explicit contrast-transition models. To separate the effect of what is remembered from how long it is retained, Section~4.1 also compares decayed OR-TS with discounted Beta-TS at the same decay rate (Raj and Kalyani 2017).

\engsec{Empirical evidence}

We first test Section~3's predictions in synthetic environments, separating the effects of state representation and forgetting (Section~4.1). We then examine whether the corresponding baseline and contrast variation appears in real-world experiments (Section~4.2). Finally, we compare policies in an environment estimated from the ASOS experiments (Section~4.3).

\engsub{Synthetic simulations: separating representation and forgetting}

\begin{figure*}[!t]
\centering
\includegraphics[width=\linewidth]{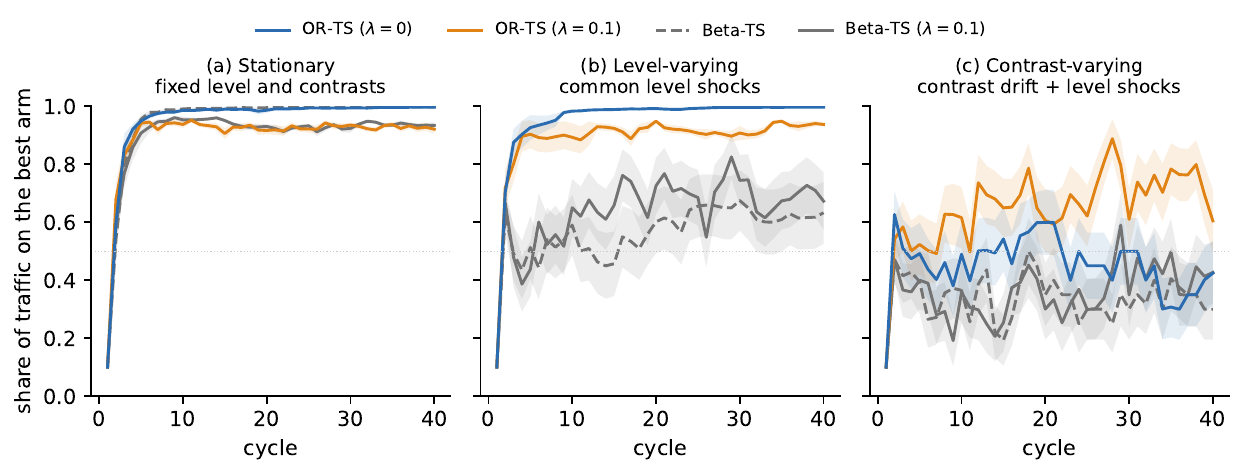}
\caption{Traffic share assigned to the true best arm in the three named synthetic environments. In the notation of Section~3: (a) Stationary, $\alpha_t=\alpha$ and $\bbeta_t=\bbeta$; (b) Level-varying, $\alpha_t$ redrawn every batch (sd $0.30$) and $\bbeta_t=\bbeta$; (c) Contrast-varying with level shocks, the same $\alpha_t$ shocks and $\bbeta_t$ a per-arm random walk (sd $0.06$ per batch). Curves show means and bands $\pm1$ se across repetitions; the horizontal dotted line marks a majority of traffic. Blue: OR-TS without decay; orange: OR-TS with decay; grey dashed: Beta-TS; grey solid: discounted Beta-TS. Higher curves indicate more traffic reaching the best arm. Cumulative regret and the contrast-drift-only controls appear in Supplement~F, Table~\ref{tab:regret}, alongside the follow-up runs.}
\label{fig:price}
\end{figure*}

The initial comparison uses fully synthetic environments: reward probabilities and disturbance processes are prespecified, not fitted to any real experiment, and each disturbance mechanism is tested on its own at a severity fixed in advance. The design is ten arms, forty batches of $100{,}000$ trials, a $3\%$ baseline rate, and effects $0.05$ apart on the logit scale. Decay settings are likewise fixed in advance, so that the effect of the state representation can be told from that of forgetting; each setting is read on its own, not as the output of a procedure that picks a setting per environment.

The common shock is a shift of the shared level redrawn every batch with sd $0.30$ on the logit scale, odds swinging by some $35\%$; the per-arm random walks move each arm's contrast by sd $0.02$ or $0.06$ per batch. The common shock is the disturbance the contrast bet allows; the per-arm random walks are the ones it excludes.

Because forgetting is the obvious remedy for drift, each family is also run with it at the same rate, $\lambda=0.1$, written in parentheses after the policy name: OR-TS($0.1$) with contrast decay, and Beta-TS($0.1$), a \emph{discounted} Beta-TS (Raj and Kalyani 2017) built by applying the tempering operator of Section~5.1 to the Beta state, which in that method's notation is a discount factor of $0.9$ per batch, so that the representation is separated from the control. This comparison gives both representations the same evidence-retention rate; recent-observation weighting is also used in platform practice (Optimizely 2026a).

Figure~\ref{fig:price} draws the three environments of Section~3, one per panel; here the Contrast-varying environment carries the common level shock as well, and contrast-only drift, also Contrast-varying, is reported in Supplement~F. Cumulative regret is the conventional report and is tabulated for the same runs in Table~\ref{tab:regret}; the figure shows instead the share of traffic on the best arm: the allocation split a dashboard displays, read against the arm the simulation knows to be best, and regret is what that share leaves on the table each cycle. The horizontal axis is the allocation cycle and the vertical axis is the traffic share assigned to that cycle's true best arm, so higher is better. The best arm is redetermined every cycle from the true rates: with level shocks alone it never changes, but under contrast drift it can, so in the Contrast-varying environment regret is measured against a moving oracle, no policy can reach zero, and only the differences between policies are read.

\runin{Stationary (panel a).} Beta-TS and OR-TS differ by less than $0.1$ se in cumulative regret: fitting a fresh level costs nothing detectable when retaining the level would have been valid, in this design.

Forgetting is not free here. Both discounted variants settle near $93\%$ of traffic on the best arm instead of $99\%$, roughly doubling the undisturbed regret ($2{,}226$ against $1{,}152$ for OR-TS, $2{,}161$ against $1{,}157$ for Beta-TS). The registered prediction had put that premium below $1{,}000$; it came out at $1{,}074$ and $1{,}005$ and is reported as exceeded.

\runin{Level-varying (panel b).} The common level takes the sd $0.30$ shock above while the contrasts stay fixed; the panel re-runs the design of Kim and Kim (2020) under the registered protocol, and remembering the contrasts pays. After forty cycles OR-TS has the best arm carrying $99.7\%$ of the traffic, with no run below a majority; Beta-TS has it carrying $63\%$ on average, and in 7 of 20 runs the best arm is not even the majority (a descriptive summary of the same run, not a registered criterion).

In regret Beta-TS accumulates five times OR-TS's, and forgetting does not rescue it: the discounted variant lands within $0.1$ se of the plain one, because the shock is redrawn every cycle and the tested discount cannot track it. OR-TS's regret is close to the undisturbed case ($1{,}246$ against $1{,}152$); this is an empirical comparison, not invariance to arbitrary shifts.

\runin{Contrast-varying with level shocks (panel c).} Per-arm contrast drift of sd $0.06$ per batch is added on top of the common shock. At $100{,}000$ trials per batch, only OR-TS with decay ends with a majority of traffic on the best arm, $60.5\%$ against $30.0\%$ for Beta-TS, $42.7\%$ for discounted Beta-TS, and $42.4\%$ for OR-TS without decay.

Its regret is about the same with and without the added shock ($8{,}183$ against $7{,}990$), while Beta-TS($0.1$) reaches $28{,}895$; even without decay OR-TS beats Beta-TS, $21{,}342$ against $35{,}386$. At these settings, contrast decay and refitting the common level address different sources of change.

\runin{Controls kept in Supplement~F.} Three further comparisons are reported there rather than drawn. Full-TS, the same logistic model with Beta-TS's memory of the level, does no better than Beta-TS under the common shock, so it is the memory rule and not the model that earns the result. A registered sweep of the shock size leaves Beta-TS's end state intact up to shocks of twice the effect gap and collapsed at six, while under fixed $1/K$ allocation Beta-TS identifies the best arm at every shock size: the level hurts it through the allocation imbalance the bandit itself creates, as Section~3 predicted (Figure~\ref{fig:tolerance}). Where only the contrasts move, the forgetting variants lead at the larger drift and OR-TS($0$) ties Beta-TS (Table~\ref{tab:regret}).

Taken together: at the tested settings, refitting the level preserves the usefulness of accumulated contrast evidence in the level-varying environment, which is not robustness to the contrast-varying one; in the latter, forgetting helps both families, and with the common shock added OR-TS with decay is the best of the four fixed settings, while unnecessary forgetting costs regret in the Stationary control. The level-smoothing alternatives of Section~6.2 were not compared here, and the earlier simulations in Kim and Kim (2020) remain historical illustrations. The focus of these comparisons is what to carry across batches: contrast evidence rather than absolute reward rates. Memory decay and allocation aggressiveness are the operational choices of Section~5; Supplement~F evaluates batch pooling, learning the decay from data, and a permanent level step. These supporting analyses examine their consequences rather than establish a jointly optimal setting.
\engsub{Baseline and contrast variation in real-world experiments}

\runin{Service and experimental unit: ASOS.} ASOS is an online fashion retailer. Its Digital Experiments Dataset records controlled experiments on its digital service, comparing a control experience with treatment variants (Liu et~al.\ 2021). Here an arm is an experimental variant, rather than a product being recommended. We use the released binary metric, metric~1; its business meaning is withheld, so an event cannot be identified specifically as a purchase or a click. The release provides aggregate checkpoints rather than individual shopping histories. In Section~3's terms, the question here is which environment the observed ASOS traffic resembles: Level-varying, where OR-TS's bet holds, or Contrast-varying, where it fails.

\runin{Measuring variation across experiments.} Differencing consecutive checkpoints gives each period's exposures and events, from which we estimate the control log odds $\alpha_t$ and the treatment-minus-control log odds $\beta_t$. Each estimate moves for two reasons, real change and sampling noise, so we subtract the average sampling variance from the variance across periods; what remains, the \emph{excess} standard deviation, is the movement beyond sampling noise, truncated at zero. The data, the metric, the series selection rule and the test with its failure criterion were registered before these series were computed; detection floors, sample sizes and provenance are in Supplement~E.

Across the 86 series the level's excess sd has median $0.34$ (quartiles $0.23$--$0.46$) against $0.013$ ($0.004$--$0.034$) for the contrast (Figure~\ref{fig:plane}A), and no series reverses that ordering: the registered ratio test passed with a median ratio of $25$ against a failure threshold of $3$. The contrasts are not frozen, since 49 of the 86 contrast excesses clear the floor below which an estimate cannot be told from zero, which every level excess clears. These are dispersions across the observed periods rather than the size of changes between consecutive ones, and they support examining contrast-based memory without establishing constant contrasts or an optimal decay rate.

The longest series in the dataset, the point highlighted in panel A, shows the same ordering period by period: 228 twelve-hour periods of a control and one treatment variant (Figure~\ref{fig:asos}B). The upper plot shows the control arm's log odds in each period, the level $\alpha_t$; the lower plot shows the treatment's log odds minus the control's, the contrast $\beta_t$, with pointwise uncertainty intervals around its estimates. The level drifts and stays where it went; the contrast varies on a much smaller scale. The intervals describe estimation uncertainty, rather than certifying a constant contrast. Beta-TS remembers the level shown above. OR-TS remembers the contrast shown below.

A supplementary analysis of a fixed display position in the Open Bandit recommendation log finds the opposite ordering, a Contrast-varying log, but sparse clicks limit its interpretation (Supplement~E; Saito et~al.\ 2020).

\runin{What this does and does not show.} For the fixed variants in these ASOS experiments, the contrast is relatively stable compared with the level: in Section~3's terms, this traffic is Level-varying at twelve-hour batches: $\alpha_t$ moves and $\bbeta_t$ hardly does. It is not strictly fixed-contrast, since 49 of the 86 contrast excesses clear their floors, so the evidence supports investigating contrast persistence without establishing that every contrast is constant or that every fixed arm set warrants zero decay. Where diagnostics show meaningful contrast changes, the environment is Contrast-varying and decay (Section~5.1) is the candidate response.

The ASOS analysis examines how the common level and the contrasts vary in real-world service experiments. Observations collected under fixed allocation are used to diagnose this variation; policy performance under adaptive allocation is evaluated separately in Sections~4.1 and~4.3. The data come from one retail service and the observed contrasts are small, so whether the same pattern holds in other services and for larger effects requires further evaluation. Dataset provenance and detailed analyses are provided in Supplement~E.

\engsub{Policy simulation using rates estimated from ASOS experiments}

\begin{figure*}[!t]
\centering
\includegraphics[width=\linewidth]{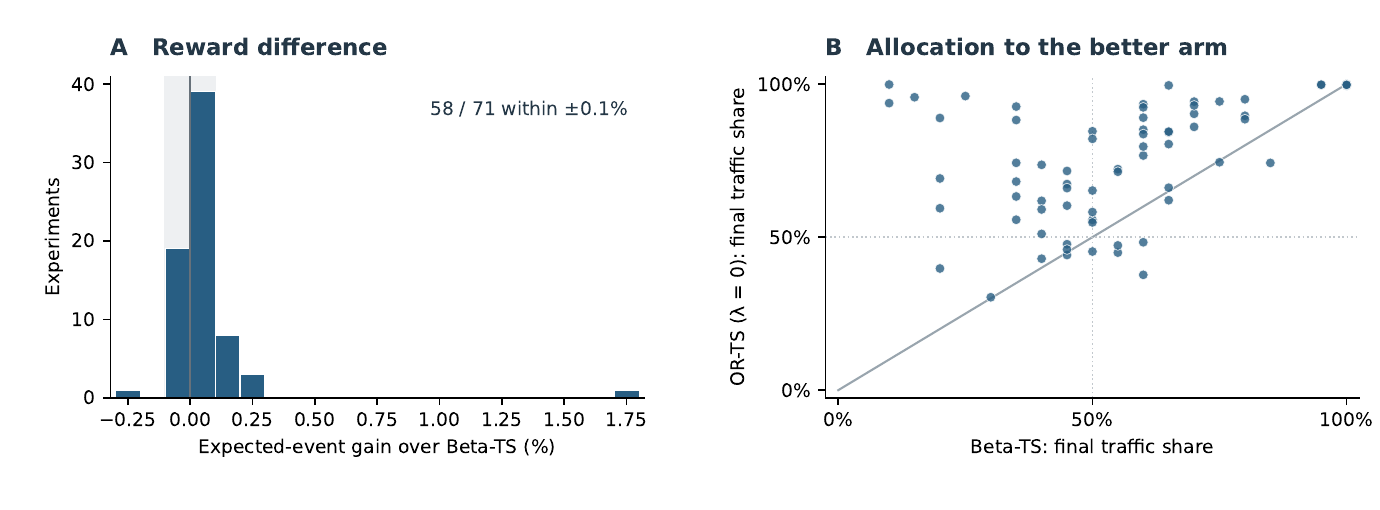}
\caption{ASOS-based policy simulation, 71 experiments, comparing OR-TS($\lambda=0$) with Beta-TS. A: the full distribution of per-experiment expected-event gain, with $0.1$ percentage-point bins and no clipping; shading marks $\pm0.1\%$. B: each experiment's final-period traffic share on the arm that is better over the whole experiment, averaged over repetitions. Points above the diagonal indicate more traffic to that arm under OR-TS; dotted lines mark a majority. The reward and allocation panels refer to the same experiments. Unlike the 86 variant--control series in Figure~\ref{fig:plane}, the unit here is an experiment. These are simulated policy outcomes using estimated rates, not outcomes of deployment at ASOS.}
\label{fig:asos_replay}
\end{figure*}

We compared the policies in simulated environments constructed from 71 real-world ASOS experiments. With small differences between arms, expected clicks were similar in most experiments, but traffic allocation differed (Figure~\ref{fig:asos_replay}). These are simulation results, not outcomes of policies deployed at ASOS. The simulations account for uncertainty in the estimated reward rates; Supplement~E gives the setup and registered outcomes.

Regret is a small fraction of the clicks, a tenth of a percent in the median experiment and under three percent in the worst, so the comparison is between small numbers. Beta-TS accumulates $4{,}848$ regret events per experiment on average against $3{,}317$ for OR-TS($0$), with Beta-TS($0.1$) and OR-TS($0.1$) in between and every Beta-TS variant above every OR-TS variant, though one three-arm experiment with a contrast of $0.025$ supplies $72\%$ of that difference. Per experiment, in expected clicks relative to Beta-TS (Figure~\ref{fig:asos_replay}, left), 58 of the 71 lie within $0.1\%$; OR-TS($0$) is ahead in 51, its worst experiment loses $0.2\%$ and its best gains $1.7\%$. The policies are nearly deterministic at these volumes, so the paired se is small and 22 experiments separate beyond two se, 16 of them against Beta-TS, although the differences are tenths of a percent. In this ASOS-based simulation, common-level variation does not translate into a benefit from contrast decay. At the tested setting of $0.1$, decay increases OR-TS regret by $376\pm49$ events per experiment and is the better setting in only 26 of the 71; discounting Beta-TS shows no resolved improvement ($-38\pm144$). These results concern the usefulness of forgetting in the estimated environment, not stationarity of its common level.

\runin{Where Beta-TS sends the traffic.} The click differences are small because the contrasts are small, not because the two policies allocate alike. Beta-TS ends with the arm that is better over the whole experiment carrying $56\%$ of the traffic on average and below a majority in 27 of the 71 experiments; OR-TS($0$) gives it $75\%$ and is below a majority in 11 (Figure~\ref{fig:asos_replay}, right). Beta-TS's cumulative averages are contaminated by the moving level through the imbalance of Section~3, so the arm they favor is only loosely tied to the better one. When Beta-TS happens to sit on the better arm, OR-TS's spread allocation costs the tenths of a percent above; when it sits on the other arm, Beta-TS pays. At these contrasts either price is small; the allocation shows that one policy finds the better arm far more often than the other, and Section~4.1 measures what that misallocation costs once the gaps are large enough to matter.

\runin{A reminder from the 2020 study.} The preprint that proposed the algorithm already reported it on eighteen days of proprietary advertising traffic, and that case is worth recalling here because its disturbance is not the one Sections~4.1 to~4.3 study. There the common level did not wander around a centre; it stepped to a new place and stayed, twice in a row and in opposite directions: clicks fell sharply on the tenth day, and on the eleventh the listings were moved to the top of the screen and page views more than doubled. Platform changes do this in practice, a page redesign, a channel switched on, a campaign started.

OR-TS earned about $3.9\%$ more expected clicks than Beta-TS and Full-TS there, most of it in the two days after the second change, while the comparators' remembered levels were freshly wrong twice over (Kim and Kim 2020). That data cannot be redistributed and the analysis carries no uncertainty interval, so it is recalled rather than reproduced.

\engsec{A Bayesian bandit agent with two controls}

A Bayesian bandit agent runs the two steps of Section~2 in a loop: recognition updates the belief from the batch that has just closed, and action turns that belief into the next batch's allocation (Figure~\ref{fig:agent}). Each step carries one control. In recognition, decay $\lambda$ sets how much of the carried evidence survives the update; in action, aggressiveness $\gamma$ sets how strongly traffic concentrates on likely winners. The allocation then produces the observations that the next recognition step consumes. OR-TS is the instance that carries the contrast posterior and fits the level afresh, with $\lambda=0$ and $\gamma=1$ as its basic setting; Beta-TS carries beliefs about absolute reward rates instead. The following subsections take the controls one at a time, and Algorithm~2 in Supplement~C specifies OR-TS's complete update.

\engsub{Posterior decay}

Decay is the control in recognition. Before fitting a batch, \emph{temper} the carried posterior, in OR-TS the Gaussian contrast posterior: raise it to a power below one, which flattens it. This applies the power-prior idea of discounting historical information (Ibrahim et al.\ 2015) to the contrast posterior:

\begin{equation}
S_{t-1}^{(\lambda)}=(1-\lambda)S_{t-1},\qquad 0\leq\lambda\leq 1.
\label{eq:decay_precision}
\end{equation}
Equivalently,
\begin{equation}
q_{t-1}^{(\lambda)}(\bbeta)\propto q_{t-1}(\bbeta)^{1-\lambda}.
\label{eq:tempered_prior}
\end{equation}
The mean is unchanged and the covariance is inflated. The new batch adds contrast information $J_t$, giving $S_t=(1-\lambda)S_{t-1}+J_t$ (Supplement~A). Zero decay retains all evidence; positive decay gives an approximate memory scale of $1/\lambda$ batches, describing how long past evidence remains influential rather than a cutoff for stored observations. At $\lambda=1$ no carried evidence enters the attempted fit, which must meet Algorithm~1's flat-prior first-fit condition. A successful fit replaces the old state; a rejected fit leaves the last validated state available for recovery (Algorithm~2). For the matched Beta-TS comparison, tempering multiplies $a-1$ and $b-1$ by the same factor, discounting toward the uniform prior (Raj and Kalyani 2017).

A common-level change alone does not call for contrast decay; the fresh intercept absorbs it (Section~4.1). The reason to discount that state is evidence that past contrasts have become less relevant. Even then, decay is a tool for where each batch brings enough evidence to replace what it discards: the criterion is not how much the contrasts move but how much they move relative to the information a batch carries, so a batch that carries fewer events changes what a given $\lambda$ means. Decay reduces stale evidence but also discards useful evidence, so the appropriate memory length depends on the application (Section~4.1). Decay is defined per update, so the batch sets what a given $\lambda$ means in calendar time: a weekly batch at $\lambda=0.1$ remembers about ten weeks, a daily one about ten days.

The rate can also be learned: in state-space terms $1-\lambda$ is a discount factor, which dynamic linear models estimate from the predictive likelihood or track online as a volatility (West and Harrison 1997; S\"arkk\"a and Svensson 2023; Mathys et al.\ 2014). Supplement~F evaluates one such rule, a bank of fixed-decay filters selected by cumulative Laplace evidence: it stays at zero where the contrasts do not move and learns to forget where they do, but lags the best fixed setting by more than the registered margin in three of five environments, so it is reported there as an option rather than the default. Supplement~G's paragraph relating $\lambda$ to a transition model is the place to start for a transition-based alternative; Supplement~G also discusses calibration and sensitivity.

\engsub{Allocation aggressiveness}

Aggressiveness is the control in action. Thompson sampling addresses the \emph{exploration--exploitation tradeoff} through posterior sampling: likely winners receive traffic, while uncertainty leaves opportunities for other arms (Russo et al.\ 2018). Aggressiveness lets an operator adjust that allocation toward more exploration or more exploitation without changing the posterior (Scott 2010, 2015). Raise the winner frequencies to a power and renormalize,
\begin{equation}
\pi_{i,t+1}(\gamma)=\frac{\rho_{i,t}^{\gamma}}{\sum_{j=1}^{K}\rho_{j,t}^{\gamma}},\qquad \gamma\geq 0,
\label{eq:aggressiveness}
\end{equation}
where $\rho_{i,t}$ is arm $i$'s winner frequency from step A1, with the convention $0^0=1$. Probability matching uses $\gamma=1$; $0<\gamma<1$ spreads positive shares more evenly, giving less-favored arms more exposure relative to the leaders, while $\gamma>1$ favors exploitation of current leaders. At $\gamma=0$ every weight is one, so allocation is uniform regardless of estimated winner frequencies, the allocation rule of a balanced A/B test, with posterior updating still active. This specifies allocation, not a stopping or hypothesis-testing rule. For $\gamma>0$, a zero Monte Carlo winner frequency remains zero, so guaranteed exposure requires explicit allocation floors.

\runin{Choosing $\gamma$.} This controls willingness to concentrate traffic, not the validity of the evidence. Stale beliefs, a Contrast-varying environment, call for a change in recognition, not in action. It can also be moved on a schedule: start at $\gamma=0$, an equal split while the first batches accumulate, and raise it as the posterior sharpens, the same device as the equal-randomization burn-in of response-adaptive trials (Villar et al.\ 2015). The experiments hold $\gamma=1$ throughout and evaluate no schedule; Supplement~G discusses the operational tradeoffs.

\engsec{Extensions}

The preceding sections specify a basic agent for a fixed arm set and a binary reward. Three directions follow. Section~6.1 lets the arm set change while earlier comparisons continue to count; Section~6.2 adds structure to the level, the contrasts, and the reward; Section~6.3 sets OR-TS beside adjacent methods.

\engsub{Changing arm sets}

\runin{The state belongs to a set of arms, not to an experiment.} A batch need only serve a subset of the stored arms: that subset has its own contrast posterior, a linear transformation of the stored one, and the winner probabilities inside it are unchanged. An arm with no history enters through the prior, which leaves the incumbents' pairwise posteriors alone. And a batch serving arms from two sets joins them: if one experiment compares A with B and the next compares B with C, the shared arm B is the \emph{bridge}, and the joint posterior carries a comparison of A with C that was never run directly, with the wider uncertainty the indirect route implies. A state therefore covers the arms that comparisons have connected so far, and groups whose arms never share a batch keep separate states. Supplement~B gives the transformations and a worked example.

\runin{Starting a new arm.} A brand-new arm has no learned coordinate yet. Its first allocation follows a rule: with $m$ new arms among the $|A_{t+1}|$ to be served, each new arm receives the uniform share $1/|A_{t+1}|$ and the observed arms' winner fractions are scaled into the remainder (Kim and Kim 2020). After its first observed batch the arm joins the joint state through the symmetric augmentation of Supplement~B. What the arrival costs depends on batch size: at $100{,}000$ trials a round OR-TS stays ahead when the level moves, while at thirty events an arm Beta-TS leads after an arrival either way, because refitting a level from a handful of events costs more than the moving level does (Supplement~F). OR-TS learns the newcomer's gap only from batches that serve it beside the incumbents, whereas Beta-TS compares it with remembered rates from the first batch, which is right while the level holds and wrong once it moves (Table~\ref{tab:situations}).

\runin{Designs this permits.} Planning can then be done in contrasts rather than in experiments. A batch need only serve a subset, so a challenger can be added or an arm retired at a boundary without restarting the comparison; a rotation that always retains one arm in common keeps everything it touches in a single component; and arms too numerous to serve at once can still be compared through the arms they share. These follow from the representation; none is an allocation algorithm and none is evaluated here.

\runin{Checking continued comparability.} This reuse assumes compatible arm definitions and stable relative effects under the changing audience and context. Because the same contrast can be reached through different bridge arms, sudden disagreement between routes is a diagnostic in its own right: evidence against a common shift, to be met with stronger decay or a restart at that boundary rather than with persistence.

\engsub{Richer models}

\runin{Structure in the level.} Basic OR-TS fits a fresh level each batch; carrying the fitted level forward instead is the Full-TS choice, and it can be made afresh at every boundary. Where the level repeats on a known cycle, keep one level per weekday, $\logit(p_{i,t})=\alpha_{d(t)}+\beta_i$ with $\beta_K=0$: Monday's observations update Monday's level while comparisons from every weekday inform the shared contrasts, which is seasonal-indicator regression (Hyndman and Athanasopoulos 2021, Section~7.4). It requires keeping the joint posterior of levels and contrasts, covariance included; a discarded level cannot be recovered later. More elaborate trends can follow standard time-series components (Taylor and Letham 2018), adapted to the reward likelihood.

\runin{Structure in the contrasts.} Shared covariate effects give a GLM and random effects for partial pooling a GLMM (Gelman and Hill 2007), of which the weekday levels are the simplest instance. Specifying how the contrasts evolve between batches gives a dynamic GLM: a random-walk transition adds an innovation covariance to the current one (S\"arkk\"a and Svensson 2023, Section~3.4) where decay inflates it proportionally (Supplement~G), and OR-TS's fresh level is the limiting transition for $\alpha_t$, the Level-varying model with no memory of the level at all. Arm-by-covariate interactions let the best arm depend on the user, a contextual extension. These enrich the model; none of them settles which posterior components cross a boundary.

\runin{The counts let the model be chosen later.} An update consumes only each batch's exposures and events per arm, so a service that keeps those counts keeps everything the recursion reads: a model chosen later can be fitted to the whole history and the state rebuilt from it, each piece added once a diagnostic gives a reason to. Two limits. The counts must already be recorded at the grain the later model needs, a logging decision taken in advance, and the allocation that produced them came from the earlier state, so it decides how much data each arm has.

\runin{Beyond binary rewards.} More generally, separate a persistent decision parameter $\phi$ from batch nuisance $\eta_t$:
\begin{equation}
q_t(\phi)\propto q_{t-1}(\phi)\int p(\Data_t\mid\phi,\eta_t)\,\pi_t(\eta_t)\,d\eta_t.
\label{eq:general_filter}
\end{equation}
The Bernoulli--logit choice $\phi=\bbeta$, $\eta_t=\alpha_t$ gives OR-TS. Gaussian rewards with an identity link and Gaussian priors allow an exact Gaussian update. None of these extensions is evaluated here; the paper's performance claims remain limited to the tested specification.

\engsub{Adjacent methods}

OR-TS is related to methods that separate treatment effects from a changing baseline and to methods that discount historical reward information: action-centered, semiparametric and deconfounded bandits make related separations under different assumptions (Greenewald et~al.\ 2017; Krishnamurthy et~al.\ 2018; Kim and Paik 2019), generalized-linear and nonstationary bandits supply related reward models and forgetting mechanisms (Filippi et~al.\ 2010; Garivier and Moulines 2011; Raj and Kalyani 2017), and response-adaptive clinical trials are a related allocation setting (Villar et~al.\ 2015). What sets OR-TS apart is the rule at the batch boundary: marginalize the level and carry the contrast posterior, fitting the current level from the batch's concurrent exposures rather than forecasting its trajectory as a sequential filter would; no dedicated control arm is needed, and the reference arm is a choice of coordinates (Section~2.2). Supplement~H compares the assumptions and aims of each method.

Practice has met the same nonstationarity from the estimation side. Optimizely's epoch statistics assume that time variation affects the arms of an experiment equally, partition the run into epochs of constant allocation, and estimate lift within each epoch before pooling (Optimizely 2026b): a stratified estimate that gives every epoch its own level. That is the fresh intercept applied to the estimate. OR-TS applies it to the state that allocates, while Optimizely's allocation engine underneath remains a per-arm bandit (Optimizely 2026a).

\engsec{Discussion}

For batched binary bandits in Level-varying environments, OR-TS is the candidate: carry the joint contrast posterior and refit the common level at every batch. The ranking identity explains why allocation can use that marginal posterior; the temporal model specifies when carrying it is justified. The logistic model, marginalization, and reference transformations are standard components; the contribution is their specification and evaluation as a batch-boundary memory rule, building on Kim and Kim (2020).

\runin{The case for OR-TS as the base state representation.} The contrast block determines the ranking (Section~2.2), and carrying $q_t(\bbeta)$ is the exact marginal recursion under the stated model (Supplement~A). It requires persistence of contrasts without also requiring persistence of the level. The real-world experiment series and the Level-varying simulation support that separation, the Stationary control measures the cost of relearning the level, the ASOS replay bounds what OR-TS costs where the contrasts are too small to resolve, $0.2\%$ of expected clicks at worst while Beta-TS ends with the wrong arm in the majority in 27 of 71 experiments (Section~4.3), and the Contrast-varying position-1 replay found no detectable cost to running OR-TS there, on the measured rates and on a measurement-error environment alike, where forgetting costs both families once the daily noise is removed (Supplement~E). The evidence supports the tested comparisons, not a universally best memory length or a policy that selects the best setting for each environment. The contrast-drift results assess a separate decay control; they are not evidence that the need for decay can be identified in advance. Supplement~F shows it can be identified from the data within a few batches, at a price that exceeded the registered margin against the best fixed decay in three of five environments.

\runin{A complete loop with two controls.} The design guide culminates in three choices: retain the contrasts, decide how much historical evidence to retain through $\lambda$, and decide how strongly the resulting belief should shape traffic through $\gamma$. The practical concern behind aggressiveness---hesitation to commit most traffic to an early apparent winner---belongs to the action module; concern that old evidence no longer describes the experiment belongs to recognition. Separating these choices makes the agent easier to reason about and operate, while their feedback through future data remains explicit.

\begin{table*}[!tb]
\centering\small
\setlength{\tabcolsep}{6pt}
\renewcommand{\arraystretch}{1.15}
\begin{tabular}{@{}>{\raggedright\arraybackslash}p{0.19\textwidth}>{\raggedright\arraybackslash}p{0.15\textwidth}>{\raggedright\arraybackslash}p{0.58\textwidth}@{}}
\toprule
\textbf{Expected environment} & \textbf{Candidate} & \textbf{Evidence and limits} \\
\midrule
Level-varying & OR-TS & Favoured in the common-shock simulations (Section~4.1); sparse newcomer data can reverse it (Supplement~F). \\
\addlinespace
Contrast-varying & OR-TS with decay & Decay helped under strong contrast drift, and kept the best-arm majority when the level moved too; with contrast drift alone, discounted Beta-TS is comparable (Section~4.1). \\
\addlinespace
Stationary; fixed arm set & Beta-TS or OR-TS & The two performed alike in the Stationary control, so keeping the current method is reasonable; no equivalence margin was registered (Section~4.1). \\
\addlinespace
Stationary rates; a new arm joins & Beta-TS & Remembered rates give the newcomer a comparison; Beta-TS led in the tested single-arrival design, and repeated arrivals were not tested (Supplement~F). \\
\bottomrule
\end{tabular}
\caption{Method candidates by the environment you expect, not a validated rule for selecting the best policy. OR-TS denotes no contrast decay unless stated otherwise. Batch information matters: in the sparse newcomer experiment, Beta-TS led even with common level shocks (Supplement~F). Check concurrent observations and event counts before relying on level refitting or decay; the tested settings do not establish a general threshold.}
\label{tab:situations}
\end{table*}

\runin{Environment conditions require diagnosis.} Table~\ref{tab:situations} does not assume that the environment is known before allocation begins. A preliminary randomized A/B test, or records of arms exposed concurrently within operational batches, can help assess whether contrasts vary beyond sampling noise (Section~4.2); a single comparison pooled over time cannot establish temporal stability. Such evidence informs an initial memory choice but does not guarantee future stability, so reassessment during operation remains necessary. With stable contrasts, accumulating evidence refines persistent comparisons; with changing contrasts, the agent must track which comparisons remain relevant. Decay is an option for this tracking, not an objective of making the agent itself fluctuate. Contrast variation need not change the best arm or make decay beneficial: its value depends on the variation relative to the information supplied by new batches (Section~5.1).

\runin{What to do.} Table~\ref{tab:situations} collects the recommendation by situation; the rest of this paragraph is the default. For a batched binary bandit whose traffic is Stationary or Level-varying, that is, whose contrasts $\bbeta_t$ appear sufficiently stable, Algorithm~1 offers a practical starting point: carry the contrast posterior, fit a fresh level, and use probability matching ($\gamma=1$), with the memory length chosen separately. A fixed arm set alone does not establish that stability. Start at $\lambda=0$ and turn decay on only when diagnostics provide evidence of meaningful contrast variation, the excess-sd check of Section~4.2 or disagreement between bridge routes, and only where batches carry enough events to replace what is forgotten (Section~5.1). An explicit contrast transition (Section~6.2) is the alternative. Supplement~G explains the additional assumptions needed to calibrate decay. Check the batch against the two conditions stated in Sections~1 and~2, one credible common level per batch and enough data for the posterior approximation, using Table~\ref{tab:fidelity} as an empirical guide, with additional validation for the intended arm count and allocation pattern. Very sparse batches warrant particular attention to the posterior approximation.

Five bounds remain. The evidence is from online services, so clinical time trends are the obvious next test, at batch sizes where Table~\ref{tab:fidelity} may not favour the Gaussian state; the simulations all use one forty-batch loop with one disturbance shape each; sparse adaptive allocations and larger arm sets are untested; stopping and arm-dropping rules are not evaluated; and the regret theory of the fresh-intercept recursion is open. Also open, and next in line: racing the level transitions of Section~6.2, partial pooling included, across batch sizes, a registered replay using publicly available traffic logs that capture a platform-wide change, and a dynamic model for $\bbeta_t$ that keeps the OR-TS distinction between comparison and level.

\runin{Code and data availability.}
The reference implementation is available at \url{https://github.com/sulgik/orts}. The synthetic runner and archived plot outputs are included with the project. The advertising data referenced in Section~4.3 are proprietary and cannot be redistributed; their replay is reported in Kim and Kim (2020).

\runin{AI-assisted revision.}
Generative AI tools assisted with literature search, critical review, and editing. The author determined the framing, checked every claim and derivation, and takes responsibility for the paper.

\vspace{1.4em}
\textcolor{rule}{\rule{\linewidth}{0.4pt}}
\vspace{0.6em}

{\sffamily\bfseries\footnotesize\color{lab}\MakeUppercase{References}}
\vspace{0.5em}

\small
\begin{list}{}{\setlength{\leftmargin}{1.4em}\setlength{\itemindent}{-1.4em}\setlength{\itemsep}{0.25em}\setlength{\parsep}{0pt}}
\item Adobe. Auto-Allocate: automated traffic allocation. Adobe Target documentation, accessed September 2026.
\item A. Agresti. \emph{Categorical Data Analysis}, second edition. Wiley, 2003.
\item K. J. \AA str\"om. Optimal control of Markov processes with incomplete state information. \emph{Journal of Mathematical Analysis and Applications}, 10(1):174--205, 1965.
\item S. Agrawal and N. Goyal. Thompson sampling for contextual bandits with linear payoffs. In \emph{Proceedings of the 30th International Conference on Machine Learning}, 2013.
\item C. M. Bishop. \emph{Pattern Recognition and Machine Learning}. Springer, 2006.
\item P. Auer, N. Cesa-Bianchi, and P. Fischer. Finite-time analysis of the multiarmed bandit problem. \emph{Machine Learning}, 47(2--3):235--256, 2002.
\item T. E. J. Behrens, M. W. Woolrich, M. E. Walton, and M. F. S. Rushworth. Learning the value of information in an uncertain world. \emph{Nature Neuroscience}, 10(9):1214--1221, 2007.
\item K. H. Brodersen, F. Gallusser, J. Koehler, N. Remy, and S. L. Scott. Inferring causal impact using Bayesian structural time-series models. \emph{Annals of Applied Statistics}, 9(1):247--274, 2015.
\item S. Bubeck and N. Cesa-Bianchi. Regret analysis of stochastic and nonstochastic multi-armed bandit problems. \emph{Foundations and Trends in Machine Learning}, 5(1):1--122, 2012.
\item O. Chapelle and L. Li. An empirical evaluation of Thompson sampling. In \emph{Advances in Neural Information Processing Systems}, 2011.
\item M.-H. Chen and Q.-M. Shao. Propriety of posterior distribution for dichotomous quantal response models with general link functions. \emph{Proceedings of the American Mathematical Society}, 129(1):293--302, 2001.
\item A. Deng. Objective Bayesian two sample hypothesis testing for online controlled experiments. In \emph{Proceedings of the 24th International Conference on World Wide Web Companion}, 2015.
\item A. Deng, J. Lu, and S. Chen. Continuous monitoring of A/B tests without pain: Optional stopping in Bayesian testing. In \emph{IEEE International Conference on Data Science and Advanced Analytics}, pages 243--252, 2016.
\item S. Dong, T. Ma, and B. Van Roy. On the performance of Thompson sampling on logistic bandits. arXiv:1905.04654, 2019.
\item B. Dumitrascu, K. Feng, and B. E. Engelhardt. PG-TS: Improved Thompson sampling for logistic contextual bandits. In \emph{Advances in Neural Information Processing Systems}, 2018.
\item L. Faury, M. Abeille, C. Calauzènes, and O. Fercoq. Improved optimistic algorithms for logistic bandits. In \emph{Proceedings of the 37th International Conference on Machine Learning}, 2020.
\item S. Filippi, O. Cappé, A. Garivier, and C. Szepesvári. Parametric bandits: The generalized linear case. In \emph{Advances in Neural Information Processing Systems}, 2010.
\item A. Gelman and J. Hill. \emph{Data Analysis Using Regression and Multilevel/Hierarchical Models}. Cambridge University Press, 2007.
\item A. Garivier and E. Moulines. On upper-confidence bound policies for switching bandit problems. In \emph{Algorithmic Learning Theory}, pages 174--188, 2011.
\item S. Greenland, J. M. Robins, and J. Pearl. Confounding and collapsibility in causal inference. \emph{Statistical Science}, 14(1):29--46, 1999.
\item N. Gupta, O.-C. Granmo, and A. Agrawala. Thompson sampling for dynamic multi-armed bandits. In \emph{10th International Conference on Machine Learning and Applications}, pages 484--489, 2011.
\item V. Hadad, D. A. Hirshberg, R. Zhan, S. Wager, and S. Athey. Confidence intervals for policy evaluation in adaptive experiments. \emph{Proceedings of the National Academy of Sciences}, 118(15):e2014602118, 2021.
\item K. Greenewald, A. Tewari, S. Murphy, and P. Klasnja. Action centered contextual bandits. In \emph{Advances in Neural Information Processing Systems}, 2017.
\item J. G. Ibrahim, M.-H. Chen, Y. Gwon, and F. Chen. The power prior: theory and applications. \emph{Statistics in Medicine}, 34(28):3724--3749, 2015. doi:10.1002/sim.6728.
\item Y. Jiang, W. Zhao, and V. Durkalski-Mauldin. Time-trend impact on treatment estimation in two-arm clinical trials with a binary outcome and Bayesian response adaptive randomization. \emph{Journal of Biopharmaceutical Statistics}, 30(1):69--88, 2020.
\item R. J. Hyndman and G. Athanasopoulos. \emph{Forecasting: Principles and Practice}. 3rd edition. OTexts, 2021. \url{https://otexts.com/fpp3/useful-predictors.html}.
\item R. E. Kalman. A new approach to linear filtering and prediction problems. \emph{Journal of Basic Engineering}, 82(1):35--45, 1960.
\item C. Kalkanli and A. \"Ozg\"ur. Batched Thompson sampling. In \emph{Advances in Neural Information Processing Systems}, 2021.
\item L. P. Kaelbling, M. L. Littman, and A. R. Cassandra. Planning and acting in partially observable stochastic domains. \emph{Artificial Intelligence}, 101(1--2):99--134, 1998.
\item G.-S. Kim and M. C. Paik. Contextual multi-armed bandit algorithm for semiparametric reward model. In \emph{Proceedings of the 36th International Conference on Machine Learning}, pages 3389--3397, 2019.
\item S. Kim, N. J. Morris, S. Won, and R. C. Elston. Single-marker and two-marker association tests for unphased case-control genotype data, with a power comparison. \emph{Genetic Epidemiology}, 34(1):67--77, 2010. doi:10.1002/gepi.20436.
\item S. Kim and K. Kim. Odds-ratio Thompson sampling to control for time-varying effect. arXiv:2003.01905, 2020.
\item D. C. Knill and A. Pouget. The Bayesian brain: the role of uncertainty in neural coding and computation. \emph{Trends in Neurosciences}, 27(12):712--719, 2004.
\item R. Kohavi, D. Tang, and Y. Xu. \emph{Trustworthy Online Controlled Experiments: A Practical Guide to A/B Testing}. Cambridge University Press, 2020.
\item A. Krishnamurthy, Z. S. Wu, and V. Syrgkanis. Semiparametric contextual bandits. In \emph{Proceedings of the 35th International Conference on Machine Learning}, pages 2776--2785, 2018.
\item T. L. Lai and H. Robbins. Asymptotically efficient adaptive allocation rules. \emph{Advances in Applied Mathematics}, 6(1):4--22, 1985.
\item LaunchDarkly. Experimentation statistical methodology for Bayesian experiments. Online documentation, accessed August 2026. Cited as 2026a.
\item LaunchDarkly. Creating multi-armed bandits. Online documentation, accessed September 2026. \url{https://launchdarkly.com/docs/home/multi-armed-bandits/create}. Cited as 2026b.
\item L. Li, W. Chu, J. Langford, and R. E. Schapire. A contextual-bandit approach to personalized news article recommendation. In \emph{Proceedings of the 19th International Conference on World Wide Web}, 2010.
\item T. Lancaster. The incidental parameter problem since 1948. \emph{Journal of Econometrics}, 95(2):391--413, 2000.
\item C. H. B. Liu, \^{A}. Cardoso, P. Couturier, and E. J. McCoy. Datasets for online controlled experiments. In \emph{Advances in Neural Information Processing Systems, Datasets and Benchmarks Track}, 2021. arXiv:2111.10198.
\item C. Mathys, E. I. Lomakina, J. Daunizeau, S. Iglesias, K. H. Brodersen, K. J. Friston, and K. E. Stephan. Uncertainty in perception and the Hierarchical Gaussian Filter. \emph{Frontiers in Human Neuroscience}, 8:825, 2014.
\item J. McKenzie. Multi-armed bandits and the Stitch Fix experimentation platform. Stitch Fix Technology, 2020.
\item J. Neyman and E. L. Scott. Consistent estimates based on partially consistent observations. \emph{Econometrica}, 16(1):1--32, 1948.
\item M.-H. Oh and G. Iyengar. Thompson sampling for multinomial logit contextual bandits. In \emph{Advances in Neural Information Processing Systems}, 2019.
\item Optimizely. Maximize lift with multi-armed bandit optimizations. Online documentation, accessed August 2026. Cited as 2026a.
\item Optimizely. How Optimizely Experimentation controls Simpson's paradox in experiments with Stats Accelerator enabled. Online documentation, accessed September 2026. Cited as 2026b.
\item V. Perchet, P. Rigollet, S. Chassang, and E. Snowberg. Batched bandit problems. \emph{Annals of Statistics}, 44(2):660--681, 2016.
\item W. B. Powell. A unified framework for stochastic optimization. \emph{European Journal of Operational Research}, 275(3):795--821, 2019.
\item C. Qin and D. Russo. Adaptive experimentation in the presence of exogenous nonstationary variation. arXiv:2202.09036, 2022.
\item V. Raj and S. Kalyani. Taming non-stationary bandits: A Bayesian approach. arXiv:1707.09727, 2017.
\item D. Russo, B. Van Roy, A. Kazerouni, I. Osband, and Z. Wen. A tutorial on Thompson sampling. \emph{Foundations and Trends in Machine Learning}, 11(1):1--96, 2018. \url{https://arxiv.org/abs/1707.02038}.
\item D. J. Schad, S. Vasishth, S. Hohenstein, and R. Kliegl. How to capitalize on a priori contrasts in linear (mixed) models: A tutorial. \emph{Journal of Memory and Language}, 110:104038, 2020.
\item Y. Saito, S. Aihara, M. Matsutani, and Y. Narita. Open Bandit Dataset and Pipeline: Towards realistic and reproducible off-policy evaluation. arXiv:2008.07146, 2020. Also in \emph{Advances in Neural Information Processing Systems, Datasets and Benchmarks Track}, 2021.
\item S. S\"arkk\"a and L. Svensson. \emph{Bayesian Filtering and Smoothing}. 2nd edition. Cambridge University Press, 2023.
\item S. L. Scott. A modern Bayesian look at the multi-armed bandit. \emph{Applied Stochastic Models in Business and Industry}, 26(6):639--658, 2010.
\item S. L. Scott. Multi-armed bandit experiments in the online service economy. \emph{Applied Stochastic Models in Business and Industry}, 31(1):37--45, 2015.
\item Statsig. Autotune methodology. Online documentation, accessed September 2026.
\item R. S. Sutton and A. G. Barto. \emph{Reinforcement Learning: An Introduction}, second edition. MIT Press, 2018.
\item S. J. Taylor and B. Letham. Forecasting at scale. \emph{The American Statistician}, 72(1):37--45, 2018.
\item W. R. Thompson. On the likelihood that one unknown probability exceeds another in view of the evidence of two samples. \emph{Biometrika}, 25(3/4):285--294, 1933.
\item S. S. Villar, J. Bowden, and J. Wason. Multi-armed bandit models for the optimal design of clinical trials: Benefits and challenges. \emph{Statistical Science}, 30(2):199--215, 2015.
\item S. S. Villar, J. Bowden, and J. Wason. Response-adaptive designs for binary responses: How to offer patient benefit while being robust to time trends? \emph{Pharmaceutical Statistics}, 17(2):182--197, 2018.
\item Z. Wang, T. Schaul, M. Hessel, H. van Hasselt, M. Lanctot, and N. de Freitas. Dueling network architectures for deep reinforcement learning. In \emph{Proceedings of the 33rd International Conference on Machine Learning}, 2016.
\item G. Welch and G. Bishop. An introduction to the Kalman filter. Technical Report TR 95-041, University of North Carolina at Chapel Hill, 2006.
\item M. West and J. Harrison. \emph{Bayesian Forecasting and Dynamic Models}, second edition. Springer, 1997.
\item VWO. Understanding the working of multi-armed bandit in VWO. Online documentation, accessed September 2026.
\end{list}

\clearpage
\noindent{\sffamily\bfseries\Large\color{lab}Supplementary material}\par
\vspace{0.3em}
\setcounter{figure}{0}\renewcommand{\thefigure}{S\arabic{figure}}\setcounter{table}{0}\renewcommand{\thetable}{S\arabic{table}}
\noindent{\footnotesize\color{lab}Supplements~A--C give the derivations, the reference-transformation algebra, and the full algorithm; Supplements~D--H the migration guide, dataset notes, the full simulation table, the deployment choices, and the literature comparison in detail. Sections are lettered; supplement figures and tables are numbered S1, S2, \ldots, apart from those of the main paper.}
\vspace{0.6em}

\engapp{Derivations and exact statements}

\runin{Marginal update recursion.}
Give the contrast vector one prior for the whole run and each batch its own independent intercept:
\[
\bbeta\sim\pi_{\bbeta},
\qquad
\alpha_t\stackrel{\mathrm{ind}}{\sim}\pi_{\alpha,t},
\qquad
\Data_t\mid\bbeta,\alpha_t\sim L_t(\bbeta,\alpha_t).
\]
The exact belief update for the persistent contrast is
\begin{equation}
q_t(\bbeta)
\propto q_{t-1}(\bbeta)
\int L_t(\bbeta,\alpha_t)\pi_{\alpha,t}(\alpha_t)\,d\alpha_t.
\label{eq:contrast_filter}
\end{equation}
Thus the previous contrast belief is multiplied by the current batch's evidence after the current level has been integrated out. The recursion is exact under the stated model. The initialization and edge cases follow below; the calculations here use the standard Laplace--Gaussian approximation for Bayesian logistic regression (Bishop 2006, Section~4.5.1).

\runin{Design matrix.}
Collect the batch's parameters as $\psi=(\alpha_t,\beta_1,\ldots,\beta_{K-1})^\top$ and give each arm one row: $x_i=(1,e_i^\top)^\top$ for $i<K$, where $e_i$ is the $i$th standard basis vector of $\mathbb R^{K-1}$, and $x_K=(1,0,\ldots,0)^\top$ for the baseline. Stacking the rows gives the $K\times K$ design matrix
\[
X=\begin{pmatrix} 1 & I_{K-1} \\ 1 & 0^\top \end{pmatrix},
\]
one intercept column and $K{-}1$ dummy columns, so that $\logit(p_{i,t})=x_i^\top\psi$. Note that $X$ is square and nonsingular: the batch has as many free parameters as arms, the fact stated in Section~2.2 that the batch fit imposes no restriction (the model is saturated), and $\psi\mapsto(p_{1,t},\ldots,p_{K,t})$ is the corresponding bijection. Aggregation is what keeps $X$ this small---one row per arm rather than one row per exposure---and it is exact for the binomial likelihood.

\runin{Aggregated binomial log likelihood.}
For aggregated counts under the reference-coded model of Equation~\eqref{eq:reference_model}, the log likelihood of batch $t$ is
\begin{align}
\ell_t(\alpha_t,\bbeta)
={}&\sum_{i=1}^{K-1}\bigl[c_{i,t}(\alpha_t+\beta_i) \nonumber\\
&\quad-n_{i,t}\log\{1+\exp(\alpha_t+\beta_i)\}\bigr] \nonumber\\
&+c_{K,t}\alpha_t-n_{K,t}\log\{1+\exp(\alpha_t)\}.
\label{eq:loglik}
\end{align}

\runin{The Laplace approximation and the marginal contrast block.}
Give $\alpha_t$ a flat (improper uniform) prior, $\pi_{\alpha,t}(\alpha_t)\propto 1$, take $\Normal(\mu_{t-1},\Sigma_{t-1})$ as the contrast prior, compute the joint MAP $(\widehat\alpha_t,\widehat\beta_t)$ of the log posterior built from Equation~\eqref{eq:loglik}, and evaluate the negative Hessian $H_t$ of the joint log posterior at that mode. The Laplace approximation is
\begin{equation}
(\alpha_t,\bbeta)\mid\Data_{1:t}
\approx \Normal\left((\widehat\alpha_t,\widehat\beta_t),H_t^{-1}\right).
\end{equation}
The next contrast posterior is the marginal Gaussian block
\begin{equation}
q_t(\bbeta)=\Normal(\mu_t,\Sigma_t),
\label{eq:gaussian_state}
\end{equation}
obtained by retaining the contrast entries of the mean and covariance. Marginalization uses the covariance block, not the corresponding precision block: for a Gaussian, dropping coordinates of the covariance matrix marginalizes them out, whereas dropping coordinates of the precision matrix conditions on them. The distinction matters because the intercept and contrasts are correlated after observing a batch. In design-matrix form the curvature is the familiar weighted cross-product: $H_t=X^\top W_t X+P_{t-1}$, where $W_t=\operatorname{diag}\{n_{i,t}\,\widehat p_{i,t}(1-\widehat p_{i,t})\}$ collects the binomial information weights at the mode and $P_{t-1}$ is the prior precision, zero in the intercept row and column under the flat prior and $S_{t-1}$ in the contrast block. The correlation between intercept and contrasts is visible there: the intercept column of $X$ is a column of ones, so it overlaps every arm's row.

\runin{Direct precision update.}
With the convention $\widehat\beta_{K,t}=0$, compute
\[
\widehat p_{i,t}
=\logit^{-1}(\widehat\alpha_t+\widehat\beta_{i,t}),
\qquad
w_{i,t}=n_{i,t}\widehat p_{i,t}(1-\widehat p_{i,t}),
\]
and put $\mathbf w_t=(w_{1,t},\ldots,w_{K-1,t})^\top$. Expanding $X^\top W_tX+P_{t-1}$ gives the joint precision
\[
H_t=
\begin{pmatrix}
\sum_{i=1}^K w_{i,t} & \mathbf w_t^\top\\
\mathbf w_t & S_{t-1}+\operatorname{diag}(\mathbf w_t)
\end{pmatrix}.
\]
The marginal contrast precision is the Schur complement of the intercept block, so R2 can be computed directly as
\begin{equation}
\begin{aligned}
\mu_t&=\widehat\beta_t,\\
S_t&=S_{t-1}+\operatorname{diag}(\mathbf w_t)
-\frac{\mathbf w_t\mathbf w_t^\top}
{\sum_{i=1}^K w_{i,t}}.
\end{aligned}
\label{eq:schur_precision}
\end{equation}
The subtraction is a rank-one correction, avoiding formation and inversion of the full joint covariance. Deleting the intercept row and column from $H_t$ would instead produce the conditional precision and is therefore incorrect for R2.

\runin{Sampling from the precision form.}
For A1, factor $S_t=L_tL_t^\top$. For each Monte Carlo draw, sample $z\sim\Normal(0,I)$, solve $L_t^\top x=z$, and set $\widetilde{\bbeta}=\mu_t+x$. Then $x$ has covariance $S_t^{-1}$, without that inverse being formed explicitly. The complete precision matrix, including its off-diagonal entries, must be retained because the contrasts are jointly uncertain.

\runin{The flat-prior first-batch alternative.}
An uninformative contrast prior is represented by $\mu_0=0$ and $S_0=0$; because the precision is zero, the value assigned to $\mu_0$ has no effect. This is the historical flat option, selected explicitly; the reference implementation's default for initial and for newly introduced contrasts is the symmetric proper prior of Supplement~A. If every arm in the first batch has both events and non-events, the first Laplace state has a closed form. Put $\widehat p_i=c_{i,1}/n_{i,1}$, $w_i=n_{i,1}\widehat p_i(1-\widehat p_i)$, and $\mathbf w=(w_1,\ldots,w_{K-1})^\top$. Then
\[
\begin{aligned}
\mu_{1,i}&=\logit(\widehat p_i)-\logit(\widehat p_K),\\
S_1&=\operatorname{diag}(w_1,\ldots,w_{K-1})
-\frac{\mathbf w\mathbf w^\top}{\sum_{j=1}^K w_j}.
\end{aligned}
\]
Thus the flat-prior first update requires no numerical optimization. If an arm has only events or only non-events, this shortcut has an infinite empirical logit and the flat-prior fit has no finite solution; the reference implementation then stops rather than fit, and the remedies are continued start-up traffic until a batch permits a fit, or a proper contrast prior specified before observing outcomes, whose scale must be reported.

\runin{Properness under the flat intercept prior.}
The integrated likelihood $\int L_t(\bbeta,\alpha_t)\,d\alpha_t$ decays exponentially as $\alpha_t\rightarrow-\infty$ if batch $t$ contains at least one event, and as $\alpha_t\rightarrow+\infty$ if it contains at least one non-event. With a proper Gaussian contrast prior, the marginal update of Equation~\eqref{eq:contrast_filter} is therefore proper exactly when the batch contains at least one event and at least one non-event (Chen and Shao 2001). With no events, the likelihood approaches its supremum as $\alpha_t\rightarrow-\infty$; with no non-events, it does so as $\alpha_t\rightarrow+\infty$. Under the improper flat intercept prior, the posterior is then improper and the joint MAP has no finite solution. This condition is specific to that prior choice, not to Bayesian logistic regression in general: a proper diffuse Gaussian prior for $\alpha_t$ yields a proper posterior and finite penalized mode. The basic flat-prior implementation skips a violating batch, $q_t=q_{t-1}$, as stated in Algorithm~1.

\runin{Why the ranking fact holds.}
The inverse-logit function is strictly increasing. Subtracting the common $\alpha_t$ from all linear predictors leaves their ordering unchanged. \hfill$\square$

\runin{Storage and initialization.}
For steps R1--R2, it is convenient to store the contrast posterior in precision form as $(\mu_{t-1},S_{t-1})$, where $\mu_{t-1}\in\mathbb R^{K-1}$ is the contrast mean and $S_{t-1}\in\mathbb R^{(K-1)\times(K-1)}$ is its precision matrix. Once the state is proper, it represents $\Normal(\mu_{t-1},S_{t-1}^{-1})$.

\runin{Symmetric proper initialization, the default, and the flat option.}
Before any outcomes have been observed, allocate the first batch uniformly,
\[
\pi_{i,1}=\frac{1}{K},
\]
giving each arm an expected $N_1/K$ exposures. The historical flat initialization sets $\mu_0=0$ and $S_0=0$ for the contrasts, the value of $\mu_0$ being immaterial because its precision is zero; it remains an explicit option that reproduces the earlier runs, and its scale-free appearance hides that it asserts every pairwise contrast may be unboundedly large.

The default is a reference-invariant proper initialization defined through exchangeable arm effects:
\[
z_i\overset{\mathrm{iid}}{\sim}N(0,\tau^2),\qquad
u_i=z_i-\bar z,\qquad \logit(p_{i,t})=\alpha_t+u_i.
\]
Here $\bar z=K^{-1}\sum_i z_i$ and $\sum_i u_i=0$; the prior is proper on this contrast subspace. With reference $K$, $\beta_i=u_i-u_K$ has the equivalent initialization
\[
\begin{aligned}
\mu_0&=0,\qquad \Sigma_0=\tau^2(I_{K-1}+\mathbf1\mathbf1^\top),\\
S_0&=\tau^{-2}\left(I_{K-1}-\frac{\mathbf1\mathbf1^\top}{K}\right).
\end{aligned}
\]
Every pairwise difference has variance $2\tau^2$, and every arm has prior winner probability $1/K$. Independent equal-variance priors on reference contrasts do not have this symmetry: they give reference--arm differences a different variance from differences between two non-reference arms. The correlated construction preserves symmetry under arm relabeling and reference changes.

Unlike flat initialization, this prior permits learning when individual arms have zero or complete event counts, provided the batch as a whole contains both events and non-events. It retains the fresh flat intercept and contrast-only memory. Its cost is a prespecified effect scale $\tau$, here $\sqrt2$: shrinkage toward the incumbents' average can delay learning a genuinely extreme difference, and the cost is largest when an arm arrives mid-run with no level to relearn (Supplement~F, H19 and H19c). The symmetric prior is the basic specification as of this revision; where the flat prior fits, the two are indistinguishable at registered margins (H13, H16--H18), and where the flat prior must skip, the default fits (H14). The independent proper contrast priors used in the earlier Open Bandit simulation are not this symmetric initialization; H18 reruns that design with both.

Choose the initialization and its scale before observing outcomes, rather than switching priors when separation is detected. The proper prior is used once at initialization; subsequent updates carry its posterior, with the stated decay, rather than adding the same prior each batch. Under either initialization, a batch with only one outcome is skipped because the intercept remains flat; allocation uses the previous state, or the start-up allocation if no fit has completed.

At every later update, the fresh intercept is introduced by forming the joint prior precision
\[
P_{t-1}=
\begin{pmatrix}
0 & 0^\top\\
0 & S_{t-1}
\end{pmatrix}.
\]
Thus a flat coordinate is implemented simply by a zero row and column. Under the flat option the same zero padding gives a newly introduced arm's contrast a flat prior; under the default the new coordinate enters through the augmentation of Supplement~B, whose covariance carries the retained latent center at variance $\tau^2/K$.

\runin{Boundary empirical rates.}
Let $\widehat p_{i,t}=c_{i,t}/n_{i,t}$. Under the default proper contrast prior an individual arm may lie at $\widehat p_{i,t}=0$ or $1$ without making the update improper. With a flat intercept, failure then occurs only if the entire batch has one outcome: $\sum_i c_{i,t}=0$ or $\sum_i c_{i,t}=\sum_i n_{i,t}$; the implementation skips such a batch and retains the previous state. Under the explicit flat option the first batch is stricter: $\widehat p_{i,1}=0$ or $1$ for any arm gives an infinite empirical logit and no finite saturated mode, so $0<c_{i,1}<n_{i,1}$ is required for every arm, and the same holds for an arm that enters later with a flat contrast prior; the reference implementation raises in that case rather than fitting a large finite value, and the caller may reject the batch. The properness conditions are stated above.

\runin{Ordering, not information invariance.}
The result above concerns ordering, not information invariance. The binomial information weight for an observation is $p_{i,t}(1-p_{i,t})$. An extreme common shift can make every outcome nearly zero or nearly one and thereby reduce information about $\bbeta$. The safe claim is that the ranking and persistent coordinates separate structurally from a common additive logit shift.

\engapp{Reference-transformation algebra}

Suppose a historical arm set $H$ uses reference $r$ and
\[
\bbeta^{H,r}\mid\Data\approx\Normal(\mu^{H,r},\Sigma^{H,r}).
\]
For a target subset $G\subseteq H$ and new reference $s\in G$, each target contrast is a linear combination of historical contrasts: with the convention $\beta^{H,r}_r\equiv 0$, the contrast of arm $a\in G$ against the new reference is $\beta^{G,s}_a=\theta_a-\theta_s=\beta^{H,r}_a-\beta^{H,r}_s$. Collecting these rows gives a matrix $T_{G,s\leftarrow H,r}$, each row of which has an entry $+1$ in the column of $\beta^{H,r}_a$ and $-1$ in the column of $\beta^{H,r}_s$ (a column corresponding to $r$ is simply omitted). Then
\begin{equation}
\begin{aligned}
\bbeta^{G,s}&=T\bbeta^{H,r},\\
\mu^{G,s}&=T\mu^{H,r},\\
\Sigma^{G,s}&=T\Sigma^{H,r}T^\top.
\end{aligned}
\label{eq:transform}
\end{equation}

\runin{Winner-probability invariance.}
Winner probabilities within $G$ are unchanged by re-referencing. In any single posterior draw, changing the reference subtracts the same arm score from every score, and the arg-max of a set of scores is invariant to subtracting a common constant. Hence the Monte Carlo winner frequencies computed from $\Normal(\mu^{G,s},\Sigma^{G,s})$ agree with those computed from the original state restricted to $G$. The off-diagonal entries of $\Sigma^{H,r}$ are essential to this construction: a new coefficient is the difference of two correlated old coefficients, and its variance depends on their covariance.

\runin{Operational sequence.}
A changing-arm OR-TS update follows four steps:
\begin{enumerate}\setlength{\itemsep}{0.15em}
\item \textbf{Project and re-reference.} Transform the historical mean and covariance to a retained bridge arm.
\item \textbf{Augment.} Add contrast priors for new arms. Under the default, each new arm's latent effect is drawn from the same $\Normal(0,\tau^2)$ population as the incumbents', and the latent center the contrast state does not identify is retained at variance $\tau^2/K$; the augmented covariance therefore has the incumbents' block unchanged, a common $\tau^2/K$ term shared by the new coordinates, and $\tau^2$ added on their diagonal. The incumbents' pairwise posteriors and the reference invariance of the state are preserved. Under the flat option the new coordinate is a zero row and column instead.
\item \textbf{Update.} Fit the current counts with a fresh batch intercept and the expanded contrast vector.
\item \textbf{Carry.} Store the marginal contrast posterior on the connected arm group.
\end{enumerate}

\runin{Traffic for a brand-new arm.} The rule stated in Section~6.1 is the default of Kim and Kim (2020): with $|A_{t+1}|$ arms to be served, of which $m$ new ones have no posterior yet, give each new arm the uniform share $1/|A_{t+1}|$, and scale the observed arms' Thompson proportions into the remaining $1-m/|A_{t+1}|$. A different initial allocation must be specified explicitly. The informative new-arm prior is no longer a separate option: it is the default augmentation of the operational sequence above, whose cost and benefit are measured in Supplement~F (H19, H19c).

\runin{A worked indirect comparison.} Suppose batch one runs $\{A,B,C\}$ and batch two runs $\{B,C,D\}$. Arms $A$ and $D$ are never observed together, yet both have posterior contrasts against the bridge arms $B$ and $C$, and the joint contrast posterior therefore carries a posterior for $A$ against $D$---with the wider uncertainty the indirect route implies. This is the connected-component principle of the previous paragraphs in its smallest instance.

\runin{Group-level covariance and continuous operation.}

The joint contrast posterior, and its covariance, belong to the connected component of arms that comparisons have linked, not to the platform as a whole. Independent experiment groups, whose arms never share a batch, keep separate states with no covariance between them; stacking them into one matrix gains nothing and would introduce a spurious coupling. Within a component, persisting the state across experiment boundaries is what makes operation continuous rather than a sequence of restarts, the always-on mode recommender deployments already run in (Ca\~namares et al.\ 2019): each new experiment starts from the component's current contrast posterior, adds its arms through a bridge, and leaves the posterior richer for the next.

\engapp{Full algorithm and numerical cautions}

\begin{figure*}[!t]
\centering
\begin{minipage}{0.94\linewidth}
\textcolor{rule}{\rule{\linewidth}{0.6pt}}\\[0.15em]
{\textbf{Algorithm 2.} OR-TS update with explicit caller-side recovery}\\[0.1em]
\textcolor{rule}{\rule{\linewidth}{0.4pt}}
\begin{algorithmic}[1]
\Require initial contrast prior; $\lambda\in[0,1]$; $\gamma\geq0$; sampling budget $M$; start-up rule; optional feasible floors
\State store the initial state and route the first batch by the prespecified start-up rule
\For{each completed batch $t$}
  \State receive finalized counts for batch $t$ and the next active arm set
  \If{the batch has no events or no non-events}
    \State keep the state without decay; route the next batch from that state or the start-up rule; \textbf{continue}
  \EndIf
  \State make a working copy of the state; transform the reference and augment new arms (Supplement~B)
  \State set working prior precision $S^-_t\gets(1-\lambda)S_{t-1}$
  \If{the explicit flat option, a new flat contrast, or a full reset prevents a finite first fit}
    \State reject this update; retain the validated state; notify the caller; route from that state or the start-up rule; \textbf{continue}
  \EndIf
  \State fit the batch with a fresh flat-prior intercept and working contrast prior
  \If{optimization or covariance validation fails}
    \State reject the working copy; raise an error; retain the last validated allocation if feasible for the next arm set, otherwise use the prespecified fallback; \textbf{continue}
  \EndIf
  \State marginalize by Equation~\eqref{eq:schur_precision}, using $S^-_t$ as prior precision
  \State commit the validated state with its batch identifier exactly once
  \State route the next batch using the covered-arm, new-arm, and optional-floor rules below
\EndFor
\end{algorithmic}
\textcolor{rule}{\rule{\linewidth}{0.6pt}}
\end{minipage}
\end{figure*}

Algorithm~2 separates the update from the platform's routing step. It extends Algorithm~1 with decay and aggressiveness and uses the changing-arm transformations of Supplement~B. The experiments use their registered settings; the optional floor rule below is an integration option, not an additional evaluated policy.

\runin{Routing from the last valid state.} For the next active set, query the joint posterior on the covered arms using $M$ draws and apply Equation~\eqref{eq:aggressiveness}. With $m$ uncovered arms among $K'$ active arms, reserve $1/K'$ for each uncovered arm and scale the covered arms' shares to total $(K'-m)/K'$. If no arm is covered, use the prespecified start-up allocation. A full contrast reset also returns to that start-up rule until a valid fit completes.

\runin{Optional exposure floors.} Let $u_i$ be the resulting shares and let prespecified floors $f_i\geq0$ satisfy $\sum_i f_i\leq1$. One explicit allocation map is
\[
\pi_i=\max\{f_i,c u_i\},\qquad
\sum_i\max\{f_i,c u_i\}=1,\quad c\geq0.
\]
The scalar $c$ can be found by monotone search; if the floors sum to one, use them directly. No subsequent normalization is applied. This may alter the new-arm shares as well as the covered-arm shares. With zero floors the base allocation is unchanged; nonzero floors require their own validation.

\runin{Monte Carlo budget.} Prespecify $M$ and record the random seed. Check stability of the resulting shares across sampling seeds at the intended arm count; a zero sampled winner count is not proof that an arm cannot be best. The paper supplies no universal sampling-budget guarantee.

A rejected update does not mutate the stored posterior. In Algorithm~2, $\lambda=1$ attempts a flat-prior refit; if it cannot complete, recovery retains the previously validated state. A deliberate administrative reset that discards that state instead requires the start-up allocation. Unused batch counts remain in the audit log; they are not silently merged into a later batch. The library returns early on whole-batch endpoints and, under the flat option, raises on a separated new arm; the caller owns the routing and recovery steps shown here.

\runin{Numerical cautions.}
The reference implementation stores a mean vector and precision matrix, evaluates the binomial objective with stable log-sum-exp arithmetic, uses checked BFGS optimization for the MAP, computes matching gradient and curvature from aggregated counts, and estimates winner probabilities by multivariate-normal Monte Carlo sampling. Initial contrasts carry the symmetric proper prior by default, as Algorithm~1 specifies, and new contrasts enter through its augmentation; the flat initialization and the independent $N(0,s^2)$ new-contrast prior remain explicit options that reproduce the historical runs. Under the improper flat intercept prior it skips batches with no events or no non-events because their posterior is improper and their joint MAP has no finite solution. Failed fits and invalid covariance matrices raise exceptions so a production caller can retain its last validated allocation; random seeds should be recorded.

\engapp{Migrating a running service}

Table~\ref{tab:migration} in Section~1 lists what changes; this supplement gives the details. For a team already operating Beta-TS, OR-TS changes the model-update module rather than the surrounding experimentation system: the same per-batch counts go in, the same winner probabilities come out, and variants, outcome counters, and the refresh schedule are untouched. Three things inside change---the update, the state it keeps, and how that state is persisted---and the paragraphs below take them in turn.

\runin{What the platform must supply.} The exposure unit and outcome window must be defined before the experiment runs. Each update receives finalized outcomes for its own exposure batch; delayed or revised outcomes require an explicit historical-data policy and must not silently be counted as observations from the current batch. Such a policy is not evaluated here. Effect reporting, guardrail monitoring, and stopping or shipping decisions remain separate platform responsibilities, and uniform allocation does not by itself supply an A/B test's inferential procedure.

\runin{What the migration costs.} Beta-TS updates two counters per arm. OR-TS performs a joint logistic fit, a contrast-matrix factorization, and Monte Carlo sampling. Dense matrix operations and sampling scale on the order of $K^3+MK^2$ per fit-and-allocation cycle, with additional cost depending on optimizer iterations. The reference implementation uses checked BFGS optimization; no deployment-latency benchmark is reported here. Storage grows with the square of the arm count: $2K$ counters become a $(K{-}1)$ mean vector and a $(K{-}1)\times(K{-}1)$ precision matrix, $K(K{-}1)/2+(K{-}1)$ numbers once symmetry is exploited. At the arm counts noncontextual bandits run at, tens rather than thousands, that is kilobytes.

Integration should measure convergence, latency, and memory at the intended arm counts and concurrent workload. The state and data-handling requirements below are also part of the migration.

\runin{Per-cycle counts, not cumulative counters.} A batched service already produces per-batch counts, so the input is unchanged. What changes is what may be discarded afterwards. A conjugate service can keep only running totals, because adding data to a Beta posterior gives the same answer whether it arrives in one batch or ten. OR-TS cannot: the fresh intercept is \emph{per batch}, so step R1 needs the exposures and conversions of each cycle separately. A platform that exposes only cumulative counters must difference consecutive checkpoints---the same operation Section~4.2 applies to the ASOS checkpoints---and a batch left with no events, or no non-events, is skipped, as Supplement~A's properness condition requires. Identification also becomes a logging requirement rather than an assumption: within each cycle, more than one arm must actually receive traffic, or the level cannot be told apart from the contrasts.

\runin{What to keep for recovery.} A Beta posterior can be rebuilt from cumulative totals alone. The OR-TS state $(\mu_t,S_t)$ cannot: it is a sequential fold over the per-batch counts, so recovery needs either the last saved state or the per-batch log to replay from scratch. A batched service normally keeps that log already, and with it the two are on the same footing; the one rule is that no batch enters the fold twice, so persist the state together with the identifier of the last batch it absorbed. The log is also the audit trail and the offline harness for tuning $\lambda$.

\runin{Warm-starting from the incumbent's posteriors.} A running service also holds state worth keeping: the incumbent's per-arm Beta posteriors can be carried into the contrast prior through the standard logit-normal approximation, inheriting the incumbent's belief about the \emph{contrasts} while deliberately discarding its belief about the level---which is the point of the migration. Supplement~G gives the two-line construction, including the shared-reference covariance it must carry. The conversion is appropriate only if the incumbent's contrasts are credible; discarding the level coordinate does not remove historical distortion from those contrasts. Supplement~G describes that qualification and the alternatives of rebuilding from batch history or starting with a prespecified fresh prior.

\runin{Validating the swap.} With those three in hand, the two services speak the same interface, so they can be run in shadow: feed both the same counters and compare the winner probabilities they emit. Under stationary traffic the rankings should agree closely---Section~4.1's undisturbed environment has OR-TS and Beta-TS at parity in final best-arm share---but the winner probabilities need not coincide batch by batch, because the two priors and posterior calculations differ (Section~3). A large gap that persists through a quiet period points to a wiring error, such as cumulative counters fed as per-cycle counts; a modest, fluctuating one is a modeling difference. Divergence is expected, and intended, exactly when the common level moves.

\engapp{Data notes for Section~4.2}

\begin{figure}[!b]
\centering
\includegraphics[width=0.92\columnwidth]{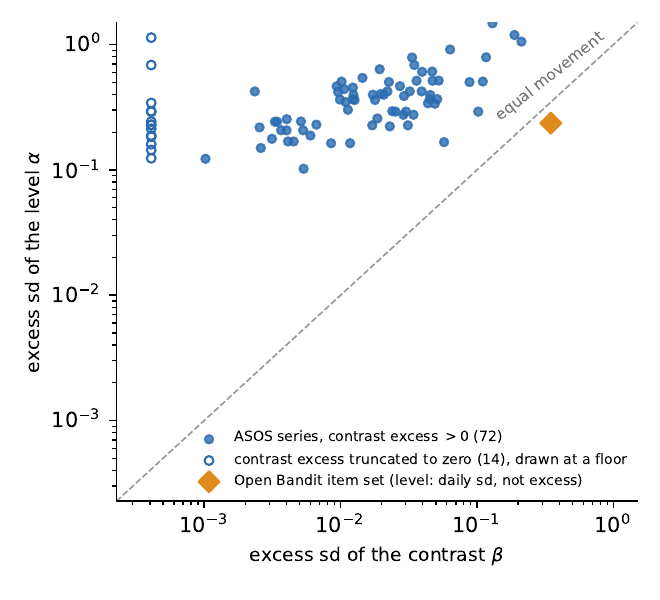}
\caption{The two excess sds paired per series: contrast (horizontal) against level (vertical), one point per ASOS series, with the equal-movement diagonal. Every series lies above the diagonal, which is the registered ratio test in one picture. The 14 series whose contrast excess was truncated to zero are open circles at a display floor, not measured positions; the Open Bandit position-1 diamond's level coordinate is the plain daily sd, not an excess.}
\label{fig:scatter}
\end{figure}

ASOS reports four anonymized metrics as group-level aggregates, cumulative from experiment start, at daily or twelve-hourly resolution (Liu et~al.\ 2021, Section~5); metric~1, the binary one, is used. Its authors count 99 variant-versus-control comparisons, released the dataset to support research on adaptive stopping rather than the use made of it here, and report that only about a quarter of the experiments reach $p<0.05$, attributing this to experiments being run where an effect is already expected---so the small contrasts are a property of the domain. They caution that the data do not represent ASOS's business, and no such conclusion is drawn here. The Open Bandit follow-up uses the ALL/random bucket at position 1. Its observed exposure counts are retained directly; no propensity weighting or position correction is used. The public aggregate is grouped by day, item, and position, and the position filter precedes every fit and rate calculation. The selected data contain $458{,}005$ displays and $1{,}622$ clicks. Its target remains the marginal rate under the logged audience and slate context, so random item exposure does not by itself rule out temporal changes in that context. Both aggregates are redistributed with the code.

\runin{ASOS policy simulation setup.} Each experiment retains its control and surviving variants, with two to four arms. Each twelve-hourly period is one batch at its actual volume: 71 experiments, $4{,}379$ batches, and twenty repetitions. Each repetition draws reward probabilities from the measurement-error environment described below before policies allocate traffic and receive newly simulated binomial outcomes. The original randomized observations supply the rate estimates; the reported simulation uses posterior draws rather than treating those estimates as fixed reward probabilities.

\runin{ASOS replay provenance.} Section~4.3 reports \path{20260915T134109Z_asos_replay_shrunk_05a47e1} (H25b), the replay on the measurement-error environment: for each experiment the measured log odds are fitted as a period level plus an arm effect with weights from the sampling variances $s_{i,t}^2=1/\text{events}+1/\text{non-events}$, each arm's excess residual variance $\tau_i^2$ is the residual variance minus the mean sampling variance floored at zero, each measurement is shrunk toward the fit by $\tau_i^2/(\tau_i^2+s_{i,t}^2)$, and each repetition draws its environment from the resulting posterior before any policy runs. The rule is the moment-method normal approximation of a binomial mixed model with a period level, an arm effect and a per-arm random deviation; at these event counts the two are not distinguishable. Its registered predictions and outcomes: every Beta-TS variant above every OR-TS variant in mean regret, met; at least 50 of 71 experiments within $0.1\%$, met at 58; at most 10 experiments separated beyond two paired se, failed at 22.

The registered primary replay used the measured rates themselves as the environment: \path{20260907T083417Z_asos_replay_ded2091} (H8) and its rerun under the symmetric default, \path{20260911T225934Z_asos_replay_symmetric_audit_d870ba6} (H17), which reproduces it. There the regret comparison was registered as a null, since at a median absolute contrast of $0.0032$ the arms are nearly interchangeable, and it was falsified: five of the six pairwise differences exceeded $2$ se and the cross-family comparisons favored OR-TS. But one experiment supplied $97.5\%$ of the Beta-TS versus OR-TS($0$) difference, the median experiment went the other way, OR-TS($0$) was ahead in only 33 of the 71 while 57 lay within $0.1\%$ (worst $-0.6\%$, best $+1.5\%$), and 31 experiments were separated beyond two se, 17 of them against OR-TS. The measured best arm of a period is itself mostly sampling noise, which is what motivated H25b. \path{20260915T134105Z_asos_replay_measured_alloc_05a47e1} (H25a) is H17 rerun with every arm's allocation recorded, identical row for row on the shared columns; on the measured rates Beta-TS ends below a majority on the better arm in 34 of 71 experiments and OR-TS($0$) in 11. H8's second registered quantity, the one-period-ahead error of each policy's belief about the control arm's log odds, with Beta-TS's cumulative posterior mean and OR-TS's just-fitted intercept as the forecasts: median absolute error $0.364$ for Beta-TS against $0.150$ for OR-TS($0$), Beta-TS the worse of the two in $88.7\%$ of the experiments with a median ratio of $2.78$, clearing the registered $70\%$ and $2$; Beta-TS($0.1$) $0.259$. The H17 rerun reproduces it ($85.9\%$, median ratio $2.82$). It is a diagnostic of the carried level's staleness, not of allocation, and the intercept is not used in OR-TS's next allocation.

\runin{Service and action: Open Bandit.} The Open Bandit Dataset comes from ZOZOTOWN, a Japanese fashion e-commerce service operated by ZOZO (Saito et~al.\ 2020). Its interface displays items in three positions and logs each displayed item's click. An arm here is an item and the event is a click, not a purchase. We fix position 1, chosen by its label before its results were seen (Supplement~E), and use only that slot in the uniformly randomized \emph{all} campaign: each row is a day--item pair, exposures count displays in that slot and events count their clicks, and no position correction or pooling enters the diagnostic or the simulation. All 80 items are observed on all seven days. The target is an item's click probability in that slot under the logged audience and the other items shown beside it, not a policy for the whole recommendation slate.

\runin{Where the contrast moves too.} At position 1 the contrasts move more than the level: pooled excess contrast sd $0.347$ against a daily level sd of $0.236$, a ratio of $0.68$ (Figure~\ref{fig:scatter}), the direction registered in advance and the reverse of ASOS. The difference is in what an arm is. An ASOS arm is an experiment variant, fixed for the duration of the test; an Open Bandit arm is one of 80 items whose relative appeal changes within the week, whether through the items themselves or through the audience and the slate around them, which the log cannot separate. That is arm-specific drift, exactly what the contrast bet excludes. The level kept moving here as well, so on this dataset both bets fail; nowhere measured did the fixed-rate bet hold. Only about three clicks per item-day support these estimates, so the contrast excess is a shrinkage estimate under the $N(0,2^2)$ prior of Supplement~E rather than a raw count; the earlier position-adjusted analysis is preserved there.

\runin{Single-position Open Bandit simulation.} The Open Bandit log (Saito et~al.\ 2020) showed 80 items uniformly at random for seven days, so each day's click rate per item at position 1 is an unbiased estimate of that day's truth, and the diagnostic above found those item contrasts moving more than the level within the week, the case where OR-TS's bet fails. The seven daily rate vectors at position 1 are taken as the environment, and the same four policies are run through the week at that slot's own volumes, about a $0.35\%$ event rate and three clicks per item-day, twenty times; this is a simulation calibrated to real daily rates, not a live comparison. Single-slot counts include item-days with no clicks, for which the flat prior of Algorithm~1 has no finite first fit, so both OR-TS variants start from a proper $N(0,2^2)$ contrast prior, fixed before the run, while Beta-TS keeps its $\mathrm{Beta}(1,1)$; this prior difference is part of the comparison.

The registered prediction was that no pair separates at this horizon, and it held: all six pairwise differences in final regret are within $2$ se, the largest being OR-TS($0$) ahead of Beta-TS by $93$ clicks in a week's total of about $2{,}300$ ($0.8$ se).

Regret is measured against each day's observed best item: the daily rates are unbiased but, at three clicks per item-day, each is a high-variance estimate, so the largest of 80 is usually a lucky one and the regret against it is inflated; only the differences are read.

At this sparsity a week is too short for any policy: on the final day the best item carries $3$--$16\%$ of the traffic, and Beta-TS ends all twenty runs with it below a majority. OR-TS's regret also varies more across repetitions than Beta-TS's (sd $500$ against $210$): a wide prior on sparse counts spreads the draws, and with them the allocation.

Running OR-TS where its bet fails cost nothing detectable. Forgetting did not cost detectably either, but at three clicks per item-day nothing has accumulated that forgetting could trade for new evidence, so that is a weak null; the earlier pooled replay at eight clicks per item-day, where forgetting did cost, is kept in Supplement~E. A measurement-error environment gives the same null between the families and removes the second: \path{20260915T142538Z_obd_position1_shrunk_990f885} (H26) applies the construction of the ASOS sensitivity run here, with half-count log odds for the zero-click item-days and one excess variance shared across items ($0.074$, median shrinkage factor $0.16$), drawing one environment per repetition. Its registered predictions and outcomes: Beta-TS minus OR-TS($0$) within two paired se, met at $+96\pm81$; every policy's regret at least $30\%$ below its measured-environment value, met at $36$--$42\%$; each forgetting variant within two paired se of its no-decay twin, failed, since decay costs OR-TS $303\pm88$ and discounting costs Beta-TS $166\pm77$. Once the sampling noise is taken out of the daily best item, forgetting is no longer free here either. The drawn best item is the item with the highest fitted effect on only $24\%$ of days, so the week still reads as Contrast-varying.

\runin{Earlier Open Bandit analyses and the single-slot follow-up.} H3 used an additive item-plus-position logistic model on all positions; its item-position likelihood was not saturated. H7 instead pooled position counts into daily item rates. These are different targets, and neither supplies the single-slot environment used by H12. Their immutable runs and summaries remain available as \path{20260830T084633Z_obd_contrast_stability_cd40101} and the original H7 run.

H12 was specified after seeing those pooled results, so it is a design correction using the same source, not an independent replication. It fixes position 1, keeps every item and day, and fits daily intercepts and reference-item contrasts without position coefficients. Item contrasts have independent $N(0,2^2)$ regularization and the intercept is unpenalized. The fixed- versus varying-contrast likelihood difference is descriptive because the fits are penalized; no chi-square rejection is claimed.

The diagnostic run is \path{20260909T215447Z_obd_position1_stability_7e61726}: daily level sd $0.236$, pooled excess contrast sd $0.347$, likelihood difference $772$ on $474$ degrees of freedom between the fixed- and free-contrast fits (descriptive), and a mean item trend of $+0.024$ per day ($t=1.4$).

The supplementary measured-environment replay is \path{20260909T215454Z_obd_position1_replay_7e61726}; the earlier pooled replay (H7), \path{20260831T123938Z_obd_replay_9a33b2b}, ran at about eight clicks per item-day and found plain Beta-TS ahead of its discounted variant by $381$ clicks ($2.1$ se), a separation that is absent at position-1 sparsity. The unchanged no-drift control was rerun alongside as a pipeline check, \path{20260909T215455Z_no_drift_control_7e61726_position1_check}, and reproduced the control values of Supplement~F exactly.

\runin{The registered secondary comparison: which scale is stable?} The contrast bet is stated on the logit scale, so that a fixed contrast is a fixed odds ratio; it could equally have been stated for the rate difference, and the two cannot both stay fixed when the level moves. The H4 registration therefore included the identical computation on the linear scale, with its own delta-method sampling variance, to be reported whichever way it came out. It came out uninformative: median $R$ of $25.12$ on the logit scale against $25.17$ on the linear scale, with the logit value larger in $56\%$ of series. At a median absolute contrast of $0.0032$ the two scales give the same number to the precision the data offer, so this dataset cannot discriminate between them. The registration admitted that outcome in advance; answering the question needs experiments with substantial effects, and it stays open.

The sampling variance of a log odds is taken as $1/\text{events}+1/\text{non-events}$, and that of a contrast as the sum for its two arms; this is the quantity subtracted from the variance across periods to form the excess variance of the main paper's Section~4.2. Under a constant quantity that estimate has sd about $\sigma_b^2\sqrt{2/(T-1)}$ for per-period sampling sd $\sigma_b$ and $T$ periods, so an excess sd below $\sigma_b\,(8/(T-1))^{1/4}$---two such sds, square-rooted---is not distinguishable from zero. At the median series length of 62 periods the factor is $0.60$; with median sampling sds of $0.013$ for the level and $0.018$ for the contrast, the floors are $0.008$ and $0.011$. All 86 level excesses exceed their floor and 49 of the 86 contrast excesses do. Across series the level's lag-one autocorrelation is $0.48$, indicating temporal persistence, and the contrast's is $-0.03$, compatible with estimation noise but also with changes lacking serial dependence. The series number 86, from 71 experiments, with 5{,}516 twelve-hourly periods in all and a median of 33{,}285 control trials per period. The registered ASOS statistic was the ratio $R$ of the two excess sds, with failure defined as a median below $3$ or fewer than $70\%$ of series above $1$; it came out at a median of $25.1$ with all 86 series above $1$ (Figure~\ref{fig:scatter}). Negative excess estimates are truncated to zero; for the contrast this happened in 14 of the 86 series, which $R$ then reports as infinite. Over the 72 finite ratios the median is $19.2$ with quartiles $10.5$ and $37.3$; $R$ exceeds $3$ in 84 series. Among the thirteen public online-controlled-experiment datasets catalogued by Liu et~al.\ (2021), only ASOS and the Udacity free-trial screener carry checkpoints within an experiment, and the screener's treatment acts after the one metric with enough traffic to track, so its contrast is null by construction. No public dataset was found in which the level and the contrasts of a fixed arm set both hold still; that is why the case of zero movement on both axes of Figure~\ref{fig:plane} is priced by simulation alone.

\begin{figure*}[t]
\centering
\includegraphics[width=0.96\linewidth]{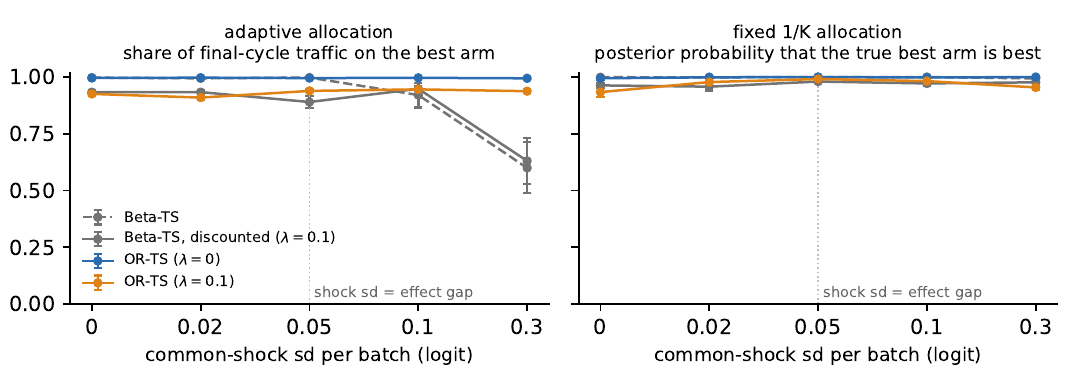}
\caption{Where Beta-TS's tolerance for a moving level ends (H11). The common-shock sd is swept from $0$ to $0.3$ against an effect gap of $0.05$ between adjacent arms, same loop as Table~\ref{tab:regret}; means $\pm1$ se over 20 repetitions. Left, adaptive allocation: the share of the final cycle's traffic on the true best arm. Beta-TS stays at parity with OR-TS through one gap, is not detectably lower at two ($0.92$ against $1.00$, $1.4$ se, with two runs in twenty below a majority) and collapses at six ($0.60$ against $0.99$, $3.5$ se). Right, fixed $1/K$ allocation: each policy's posterior probability after the last update that the true best arm is best. Beta-TS identifies the best arm as well as OR-TS at every tested shock size under fixed allocation, linking the adaptive-allocation loss to unequal exposure across changing levels. The registered boundary of one gap was too strict and is reported as such.}
\label{fig:tolerance}
\end{figure*}

\engapp{Simulation results in full}

\begin{table*}[!t]
\centering
\begin{tabular}{@{}lrrrr@{}}
\toprule
\textbf{Disturbance} & \multicolumn{2}{c}{\textbf{Beta-TS}} & \multicolumn{2}{c}{\textbf{OR-TS}} \\
\cmidrule(lr){2-3}\cmidrule(l){4-5}
& $\lambda=0$ & $\lambda=0.1$ & $\lambda=0$ & $\lambda=0.1$ \\
\midrule
Stationary
& $1{,}157\pm 35$ & $2{,}161\pm 42$ & $1{,}152\pm 50$ & $2{,}226\pm 58$ \\
Level-varying: common, sd $0.30$
& $6{,}317\pm 1{,}222$ & $6{,}462\pm 975$ & $1{,}246\pm 54$ & $2{,}476\pm 122$ \\
Contrast-varying: per-arm, sd $0.02$
& $3{,}748\pm 1{,}005$ & $3{,}051\pm 294$ & $3{,}710\pm 759$ & $3{,}090\pm 263$ \\
Contrast-varying: per-arm, sd $0.06$
& $20{,}276\pm 4{,}532$ & $9{,}911\pm 1{,}719$ & $18{,}218\pm 4{,}538$ & $7{,}990\pm 1{,}298$ \\
Contrast-varying: both ($0.30$, $0.06$)
& $35{,}386\pm 5{,}608$ & $28{,}895\pm 3{,}968$ & $21{,}342\pm 3{,}364$ & $8{,}183\pm 905$ \\
\bottomrule
\end{tabular}
\caption{Final cumulative regret, mean $\pm$ se over 20 repetitions: ten arms, forty batches, $100{,}000$ trials per batch, baseline rate $3\%$, effects $k\cdot0.05$ on the logit scale, and probability matching ($\gamma=1$). Common shocks are redrawn each batch; per-arm disturbances are random-walk increments, with sd given in logit units. Run identifiers are in Supplement~F. Decay $0.1$ retains $0.9$ of prior evidence per update. No significance marking is applied.}
\label{tab:regret}

\end{table*}

Table~\ref{tab:regret} collects the final cumulative regret of the five disturbance comparisons summarized in Figure~\ref{fig:price} and Section~4.1: the same runs, the same twenty repetitions, one row per environment; the rows come from H9 (Stationary), H6 (Level-varying and contrast-only Contrast-varying), and H10 (Contrast-varying with level shocks). The shock sweep and implementation-correction reruns are reported separately below; the ASOS comparison is in Section~4.3. Two of its rows are not drawn in Figure~\ref{fig:price}: where only the contrasts move, at sd $0.02$ per batch every policy is within about one se of every other, and at $0.06$ the two forgetting variants lead while OR-TS($0$) against Beta-TS is a tie. Full logistic Thompson sampling (Full-TS), which fits the same reference-coded model as OR-TS but carries the joint posterior of a stationary level and the contrasts, is the control that separates the model from the memory rule; its common-shock regret is reported with the reruns below and is no better than Beta-TS's. Labels H1--H12 are those of the pre-registration record, \path{docs/RESEARCH_PLAN.md} in the public code repository (\url{https://github.com/sulgik/orts}); each run id below names the immutable results directory that run produced.

\runin{Where Beta-TS's tolerance ends (H11).} The sweep of Figure~\ref{fig:tolerance} was registered with two predictions: under adaptive allocation, OR-TS($0$) and Beta-TS at parity in final best-arm share at shock sd $0$ and $0.02$ and OR-TS ahead by more than 2 se at $0.1$ and $0.3$; under fixed allocation, the two within 2 se in the posterior probability of the true best arm at every sd. The second held at every sd (largest difference $1.3$ se). The first held at $0$, $0.02$, $0.05$ and $0.3$ ($3.5$ se) and failed at $0.1$ ($1.4$ se): the boundary of one effect gap was too strict, and the break lies between two and six gaps at this batch size. OR-TS($0$)'s share is $0.994$--$0.997$ across the sweep; the discounted variants sit at $0.89$--$0.95$ as in the undisturbed environment and are not hurt at $0.1$; at $0.3$ Beta-TS($0.1$) falls to $0.63$ while OR-TS($0.1$) holds $0.94$. The fixed-allocation result is the mechanism test: each Beta-TS arm posterior is a cumulative average, so a common shift distorts the apparent contrast only insofar as the arms mix old and new evidence in different proportions, which an uneven allocation produces and a fixed one does not; OR-TS's contrast posterior never averages the level in, so the same imbalance leaves its comparison undistorted.

\runin{Implementation-correction reruns.} With the stable likelihood and endpoint convention of the reference implementation, and flat initial contrasts as Algorithm~1 specifies, H1 was rerun without changing its seed, repetitions, or horizon. Under the common shock, final cumulative regret was $1{,}170.13\pm52.83$ for OR-TS, $7{,}337.45\pm2{,}103.28$ for Full-TS, and $5{,}897.76\pm1{,}177.54$ for Beta-TS. In the no-drift control the corresponding values were $1{,}192.93\pm34.82$, $1{,}127.48\pm23.41$, and $1{,}158.30\pm26.50$: the large common-shock advantage disappeared, although no equivalence margin was registered. The immutable runs are \path{20260907T141525Z_drifting_baseline_875fd5c_flat} and \path{20260907T141614Z_no_drift_control_875fd5c_flat}. An intermediate rerun that gave initial contrasts a proper $N(0,2^2)$ prior, and the original runs, remain in the repository with additive supersession notes.

\runin{The symmetric default (H13--H19c).} This revision makes the symmetric proper contrast prior of Supplement~A the default, with $\tau=\sqrt2$ so that every pairwise log-odds contrast has prior SD $2$. Seven registered hypotheses across fourteen runs, seeds and horizons fixed before any outcome was read, ask what the change costs. \emph{Where the flat prior fits, the two are one policy.} On the H1 design (H13, runs \path{20260911T221857Z_symmetric_prior_drifting_baseline_d0714c3} and \path{20260911T221944Z_symmetric_prior_no_drift_control_d0714c3}) final cumulative regret is $1{,}310.6\pm68.9$ against the flat prior's $1{,}308.0\pm47.6$ under the common shock and $1{,}174.4\pm37.3$ against $1{,}155.1\pm31.4$ in the control, inside the registered $10\%$ margins, while the shock advantage over Beta-TS and Full-TS holds at $4.3$ and $2.9$ paired se. The same equivalence held on the shock sweep (H16, \path{20260911T224800Z_level_tolerance_symmetric_audit_d870ba6}), the ASOS replay (H17), the Open Bandit position-1 replay (H18, \path{20260911T224324Z_obd_position1_symmetric_audit_d870ba6}) and the two-arm fidelity benchmark (H15, \path{20260911T224447Z_laplace_fidelity_symmetric_d870ba6}), leaving Supplement~G's approximation check unchanged. \emph{Where the flat prior must skip, the default fits.} At 400 trials over ten arms (H14, \path{20260911T224315Z_sparse_events_drifting_d870ba6} and \path{20260911T224453Z_sparse_events_no_drift_d870ba6}) it fitted every batch against $90.9\%$ and $83.1\%$ for the flat prior, which skips a batch when an arm enters separated, and ended at $124.7\pm5.8$ against flat $137.1\pm4.4$ under the shock. \emph{When an arm arrives, every OR-TS variant trails Beta-TS.} With the true best arm arriving at round 20 of 60 (H19, \path{20260911T224624Z_new_arm_drifting_d870ba6} and \path{20260911T224712Z_new_arm_no_drift_d870ba6}) the shock advantage survives the arrival, Beta-TS minus the default being $7{,}054.9$ post-arrival at $5.5$ paired se, but with no level to relearn the augmentation costs: post-arrival regret $648.4$ against Beta-TS's $446.7$, resolved at $5.3$ se, and the three OR-TS augmentations are placed neither together nor apart by twenty repetitions. H19b was a degenerate design, identified from a single batch under every prior, and is recorded as falsified. H19c quadrupled the newcomer's prior variance and did not measurably speed its recognition or lower its post-arrival regret; at thirty trials an arm Beta-TS led every OR-TS variant post-arrival under the shock as well ($13.3$ against $26.3$, $4.7$ se), because refitting a level from a handful of events costs more than a moving level does.

\runin{A learned decay (H20--H22).} Section~5.1 fixes $\lambda$; H20 asked whether the data can choose it. The rule, fixed before the run, carries one symmetric-prior contrast filter per decay in $\{0,0.05,0.1,0.3\}$, scores every batch under each filter by the Laplace log-evidence of that batch given the filter's tempered prior, and allocates the next batch from the filter with the largest cumulative score; selection never touches the filters' memory. This is discount-factor selection by predictive likelihood (West and Harrison 1997). Run \path{20260912T021318Z_adaptive_decay_sweep_bde342b}, twenty repetitions over the five cells behind Table~\ref{tab:regret}. It behaves as intended at the two ends: the mean selected decay is $0.002$ where nothing moves and $0.205$ where the contrasts drift hardest, so it stays at full memory in the stationary cell and learns to forget under drift, beating $\lambda=0$ by $10{,}452$ (se $3{,}412$) where level and contrasts both move. The registered no-loss criterion, within $\max(10\%,2\,\mathrm{se})$ of the best fixed decay in every cell, failed in three of five, by $184$ (se $78$) against a $156$ margin with nothing moving and by $7{,}323$ (se $2{,}526$) against $5{,}052$ when both move. The mechanism the design can show is a lag: the cumulative score leaves zero only from about the fourth batch, so early batches are allocated from the wrong filter. On the one measured environment where the contrasts move, Open Bandit position 1 (H21, \path{20260912T054821Z_obd_position1_adaptive_decay_561fb93}), the rule selected $\lambda=0.3$ from the fourth day and finished $557$ clicks behind the full-memory variant (paired se $77$, registered margin $540$): at three clicks per item-day the widest tempered prior fits each sparse batch best by the evidence score, but allocating from it discards what little seven batches accumulate. Both of H21's predictions failed. H22 (\path{20260912T122357Z_adaptive_rules_sweep_ebbe1c1} and \path{20260912T122357Z_obd_position1_adaptive_rules_ebbe1c1}) kept the bank and changed only the selection rule---an evidence-weighted mixture, a retrospective allocation-reward score, and one decay re-estimated under a Beta(1,4) prior---and none passed the no-loss criterion in every cell. Learned decay therefore remains an implemented option (\texttt{AdaptiveDecayBandit}), not the default, on four rules and two environments; the one regularity is that a prior on the decay buys the stationary case cheaply.

\runin{Batches closed by event count (H23).} On the same Open Bandit environment, run \path{20260913T011906Z_obd_position1_batch_trigger_4f711dc} (seed 20260917, 20 repetitions) closed each batch when the policy's own pooled events reached $300$ (about five updates a week), $600$ (about three) or one event for every item (one or two), against daily batches, with the allocation held between updates and regret still scored daily. Paired against daily, the $300$ trigger was within noise for the full-memory variants ($-52$ and $+34$ clicks) and better for the decay variants ($-247$ and $-287$, $2.5$ and $3.4$ se), since fewer updates are fewer decay steps; the $600$ trigger cost the symmetric full-memory variant $359$ ($3.2$ se), inside the registered margin of $552$; the every-item trigger cost it $709$ and the historical variant $687$, outside their margins, and $330$--$430$ for the rest. The registered no-loss prediction failed, so event-count batches are reported as a measured trade-off, not a recommendation: on a log whose best item moves daily, an allocation held for three or more days trails the environment by more than the extra precision returns, and waiting for the rarest item to report means waiting most of the week.

\runin{Common-level steps, exposure controls, and scale sensitivity (H24--H24b).} The configs and predictions were recorded before their respective runs. H24 tests all combinations of a traffic factor of $1$ or $10$ with a permanent logit step of $0$, $+1$ or $-1$. H24b was registered after H24, using the full-precision ASOS median excess level sd ($0.341$ rounded) for the four nonzero-step cells. Both use seed 20260914, fixed contrasts and the current symmetric prior. Post-boundary regret covers batches 21--40. All six cells have identical pre-boundary histories within each suite; per-policy reward and posterior-sampling streams are seeded separately, and no state reset occurs. The unchanged no-drift audit is retained separately.

Table~\ref{tab:rollout_regret} reports every policy and condition. H24's paired Beta-minus-no-decay-OR gap is $3{,}983.9\pm1{,}569.4$ for expansion with an upward step and $797.6\pm87.9$ with a downward step; both exceed two paired se. At the ASOS scale only the downward direction exceeds the threshold (Section~4.1). These are prespecified two-se comparisons, not multiplicity-adjusted significance claims. The exposure-adjusted interaction is the change in the Beta-minus-OR post-regret gap per thousand trials caused by adding expansion to the level step, after subtracting the corresponding no-step change. It is $-0.139\pm0.039$ in the ASOS-scale downward cell: expansion reduces this additional loss-rate gap. The other three scale-by-direction interactions do not exceed two se in absolute value. Neither a positive mean alone nor a smaller absolute regret under a lower success rate establishes stronger adaptation.

The processed suites \path{rollout_h24_20260914} and \path{rollout_h24b_asos_20260914} contain all per-repetition endpoints and the full raw-run mapping; their \texttt{summary.json} files supply the numbers above. The raw runs, all under commit \texttt{c536b2f} with exact uncommitted-source snapshots, are listed below. No earlier run is superseded.

{\footnotesize\raggedright
\noindent\path{20260914T104133Z_rollout_none_c536b2f}\par
\noindent\path{20260914T104234Z_rollout_traffic_only_c536b2f}\par
\noindent\path{20260914T104134Z_rollout_level_up_c536b2f}\par
\noindent\path{20260914T104236Z_rollout_level_down_c536b2f}\par
\noindent\path{20260914T104135Z_rollout_both_up_c536b2f}\par
\noindent\path{20260914T104238Z_rollout_both_down_c536b2f}\par
\noindent\path{20260914T104517Z_rollout_asos_level_up_c536b2f}\par
\noindent\path{20260914T104518Z_rollout_asos_level_down_c536b2f}\par
\noindent\path{20260914T104519Z_rollout_asos_both_up_c536b2f}\par
\noindent\path{20260914T104520Z_rollout_asos_both_down_c536b2f}\par
\noindent\path{20260914T104136Z_no_drift_control_c536b2f_h24_control}\par
}

\begin{table*}[!t]
\centering\small
\begin{tabular}{@{}lrrrrr@{}}
\toprule
Level step & Traffic factor & Beta-TS($0$) & Beta-TS($0.1$) & OR-TS($0$) & OR-TS($0.1$) \\
\midrule
0 & 1 & $65.0\pm8.3$ & $372.2\pm14.7$ & $54.4\pm9.4$ & $395.9\pm15.6$ \\
0 & 10 & $230.9\pm18.5$ & $887.9\pm39.1$ & $230.4\pm29.6$ & $1{,}042.1\pm66.8$ \\
+1 & 1 & $432.4\pm157.5$ & $2{,}958.2\pm481.7$ & $97.7\pm17.6$ & $588.0\pm21.8$ \\
-1 & 1 & $118.3\pm9.9$ & $195.5\pm6.1$ & $25.9\pm4.9$ & $232.5\pm9.1$ \\
+1 & 10 & $4{,}316.3\pm1{,}570.7$ & $6{,}717.4\pm2{,}194.8$ & $332.4\pm35.2$ & $1{,}414.6\pm56.0$ \\
-1 & 10 & $925.8\pm92.5$ & $1{,}070.6\pm35.9$ & $128.2\pm16.0$ & $749.2\pm31.5$ \\
+0.341 & 1 & $204.9\pm78.8$ & $605.5\pm95.6$ & $69.0\pm12.4$ & $476.6\pm15.3$ \\
-0.341 & 1 & $165.9\pm17.5$ & $330.0\pm11.7$ & $51.2\pm10.4$ & $337.2\pm11.2$ \\
+0.341 & 10 & $1{,}719.6\pm794.4$ & $1{,}019.6\pm279.1$ & $249.8\pm26.5$ & $1{,}137.7\pm38.1$ \\
-0.341 & 10 & $938.6\pm113.1$ & $1{,}109.1\pm49.4$ & $173.7\pm22.3$ & $880.0\pm40.3$ \\
\bottomrule
\end{tabular}
\caption{H24/H24b post-boundary cumulative expected regret, mean $\pm$ se over twenty repetitions. Level steps are in logits; traffic factors apply from batch 21. The two zero-step controls are shared across scales and appear once. Positive-step and negative-step comparisons include all four policies; paired Beta-minus-OR differences are reported separately in the text. Values are generated from the processed H24/H24b summaries.}
\label{tab:rollout_regret}
\end{table*}

\engapp{Deployment choices in detail}

\runin{Warm start from Beta posteriors.} If the incumbent's per-arm posteriors are $\mathrm{Beta}(a_i,b_i)$, the standard logit-normal approximation carries them over: arm $i$'s log odds are approximately Gaussian with mean $\log(a_i/b_i)$ and variance $v_i=1/a_i+1/b_i$, so the contrast prior can be initialized as $\mu_0[i]=\log(a_i/b_i)-\log(a_K/b_K)$ with $\Sigma_0[i,i]=v_i+v_K$ and $\Sigma_0[i,j]=v_K$ off the diagonal---the shared-reference covariance that reference coding induces. This conversion inherits the incumbent's contrast beliefs; it does not repair a comparison already distorted by unequal historical allocation under a changing level. Use it only when those comparisons are credible. Otherwise rebuild the OR-TS state from available finalized batch history, keeping each batch's own intercept, or use a prespecified fresh initialization.

\runin{Choosing $\gamma$, in full.} The practical motivation is control over traffic concentration while evidence develops. Lowering $\gamma$ below one makes positive allocation shares more even without weakening the contrast posterior; raising it concentrates traffic on the current leaders, who need not be the true best arms. For $\gamma>0$, an arm with no Monte Carlo wins still receives zero under the power map, so an explicit floor is required to guarantee its exposure. Floors provide a minimum allocation; $\gamma$ controls the relative concentration above that constraint. Allocation choices also affect the data available for later contrast estimation and inference from adaptively collected observations (Hadad et al.\ 2021). These feedback effects are why the controls should be monitored together even though they act in separate modules. One schedule is worth naming because services reach for it: hold $\gamma=0$ for the first batches, so that traffic is split equally while every arm is still poorly measured, then raise $\gamma$ toward one as the posterior sharpens. It buys early exposure at a known price, the regret of serving a losing arm equally during the ramp, and it is the allocation counterpart of the equal-randomization burn-in used before adaptive randomization in clinical trials (Villar et al.\ 2015, 2018). How long to hold and how fast to raise are engineering choices that depend on the arm count and the events per batch. The present experiments hold $\gamma=1$ throughout and validate no such rule.

\runin{Choosing $\lambda$, in full.} The level moved more than the contrast in all 86 ASOS series, but only 14 had estimated excess contrast variance truncated to zero. Relative stability is evidence for the representation, not proof that $\lambda=0$ is optimal. Zero decay is a starting choice when contrasts are sufficiently stable at the decision horizon; positive decay is a candidate when diagnostics indicate meaningful contrast changes. Table~\ref{tab:regret} measures the cost of unnecessary decay and its benefit under strong simulated drift, while the short Open Bandit replay in Supplement~E does not distinguish the OR-TS settings. Tempering weights historical likelihood contributions geometrically in an exact power update; $1/\lambda$ batches is an approximate memory scale for the Gaussian implementation. The appropriate setting also depends on batch information and the timescale of change.

\runin{Relating $\lambda$ to a transition model.} In a scalar Gaussian state, adding an assumed innovation variance $W$ to posterior variance $v$ gives $v+W$ (the symbol $\tau$ is already the contrast prior's scale in Supplement~A). Matching that variance to tempering gives $\lambda=W/(v+W)$. This algebra does not identify $W$ from the excess variance in Section~4.2: that statistic measures variation of contrast levels across the observed periods, whereas innovation variance measures per-step changes. For a random walk, variation across levels depends on the observation horizon. Estimating innovation variance requires a transition model and allowance for sampling error, for example through a state-space likelihood; it is not evaluated here. In multiple dimensions, a scalar discount matches a transition covariance only when that covariance is proportional to the current posterior covariance. The reported excess-variance statistic therefore motivates checking for drift, but supplies neither an automatic discount nor a conclusion that the ASOS data imply $\lambda\approx0$. A prespecified discount with sensitivity analysis remains a practical option.

\runin{Relation to the 2020 preprint.}
OR-TS was proposed in Kim and Kim (2020); this is a standalone paper that cites that preprint, not a revision of it, and the central algorithm is unchanged. Everything else is new here: the recognition--action framing, the state-separation assumption and its measurement using data from real-world experiments, the registered replay, the migration guide, the separation of aggressiveness from decay, and the corrected treatment of overlap in changing arm sets (one bridge arm suffices, not two). Of the simulations, the common-shock environment re-runs the preprint's design under the registered protocol; the tolerance sweep, the fixed-allocation mechanism test, the contrast-drift environments, and the cost of forgetting are new. The preprint's advertising case is cited in Section~4.2 rather than reproduced, since its data cannot be redistributed.

\par\medskip
\noindent\begin{minipage}{\columnwidth}
\centering\small
\begin{tabular*}{\linewidth}{@{\extracolsep{\fill}}lrrr@{}}
\toprule
\textbf{Baseline} & \textbf{Trials/batch} & \multicolumn{2}{c}{\textbf{Winner-prob. error (pp)}} \\
\cmidrule(l){3-4}
\textbf{rate} & \textbf{(both arms)} & \textbf{Median} & \textbf{Maximum} \\
\midrule
3\% & 100{,}000 & $<0.001$ & $<0.001$ \\
 & 10{,}000 & $0.002$ & $0.021$ \\
 & 1{,}000 & $0.035$ & $0.338$ \\
 & 100 & $1.593$ & $6.244$ \\
\addlinespace[0.4em]
0.5\% & 100{,}000 & $<0.001$ & $0.003$ \\
 & 10{,}000 & $0.063$ & $1.818$ \\
 & 1{,}000 & $0.899$ & $5.580$ \\
 & 100 & $2.926$ & $8.677$ \\
\bottomrule
\end{tabular*}
\captionof{table}{Posterior winner-probability approximation error (percentage points): absolute difference between the Gaussian approximation and grid integration at the final batch. Smaller is better. Each row summarizes 10 repetitions of 40 batches, with observations split equally between two arms. The first two columns specify the baseline event rate and total observations per batch; the last two give the median and maximum across repetitions. Endpoint batches are skipped by both methods. Skip counts and numerical-reference checks are in Supplement~G.}
\label{tab:fidelity}
\end{minipage}\par\medskip

Endpoint skips across all repetitions were 24 for the 3\%, 100-trial condition, 5 for the 0.5\%, 1,000-trial condition, and 234 for the 0.5\%, 100-trial condition; all other conditions had none. Skipped updates are excluded from approximation comparisons.

\runin{Scope of the approximation check.} Because each cycle's approximate posterior becomes the next cycle's prior, Laplace error could in principle compound. The corrected pre-registered fidelity check compared the grid-integrated and approximate updates over 40 cycles, giving both the same observations from two arms under fixed 50/50 allocation. Both now apply the same endpoint-skip convention; skipped batches are counted and excluded from the approximation comparison instead of assigning the divergent flat-prior integral a finite grid-truncation value. The grid also splits the cell at $\beta=0$ evenly when computing winner probability. Median divergence decreased between the early and final checkpoints at every batch size tested. This supports accuracy in that design; it does not test feedback between approximation error and adaptive allocation, larger arm sets, or whether a refit-from-history safeguard is needed in those settings. Table~\ref{tab:fidelity} supplies no universal batch-size guarantee.

\runin{Correction provenance.} The common and rare fidelity runs are
\path{20260906T131733Z_laplace_fidelity_common_ec730ae_en6fix} and
\path{20260906T131832Z_laplace_fidelity_rare_ec730ae_en6fix}. A matched
higher-resolution check, \path{20260906T131919Z_laplace_fidelity_gridcheck_ec730ae_en6fix},
covered three large-batch trajectories and 120 cells; the maximum final-cycle
winner-probability discrepancy was $1.11\times10^{-16}$ and the maximum over
all cycles was $1.33\times10^{-6}$. It does not extend the validation to small
batches or adaptive allocation.

\runin{Interpretation and stopping.} Odds ratios are not generally \emph{collapsible} (Greenland et al.\ 1999; Agresti 2003); causal interpretation requires an explicit estimand and identification assumptions. An effect measure is collapsible when the marginal effect is a weighted average of stratum-specific effects. The odds ratio is not collapsible (Greenland et al.\ 1999), one reason it cannot be read as a universal causal effect measure without an explicit estimand. A stopping rule may use posterior winner probabilities or $\mathbb{E}[\max_j\beta_j-\beta_i]$, but the latter is loss on the contrast scale. Repeated monitoring, business costs, and inference after adaptive sampling require separate calibration (Deng et al.\ 2016; Hadad et al.\ 2021). Dropping an arm from allocation need not erase its stored contrast, but reactivation requires assessing whether its old evidence remains relevant.

\engapp{Adjacent methods in detail}

\runin{Same separation, other names.} Action-centered, semiparametric, and deconfounded bandits are the closest relatives: they too separate a treatment component from a baseline, with far weaker assumptions and correspondingly heavier machinery, and OR-TS is the strongly modeled Bernoulli special case that needs no contextual system (Greenewald et al.\ 2017; Krishnamurthy et al.\ 2018; Kim and Paik 2019; Qin and Russo 2022). The state design is semiparametric in the same sense, but along time rather than context: the contrasts are a fixed finite-dimensional parameter, while the levels $\alpha_1,\alpha_2,\ldots$ are one unrestricted nuisance per batch, growing without bound over a run---the incidental-parameter structure of fixed effects in panel data (Neyman and Scott 1948; Lancaster 2000), to which integrating the nuisance out under a flat prior is a standard response. That literature's warning applies when each nuisance parameter has few observations; here its information depends on the batch, so sparse operating regimes require direct approximation checks such as Supplement~G rather than an assumed event-count threshold. Response-adaptive clinical trials have long used logistic treatment effects with posterior allocation under patient drift (Villar et al.\ 2018; Jiang et al.\ 2020), the same reading of a familiar treatment-effect analysis as the state of a reward-maximizing batched bandit.

\runin{Same likelihood, other question.} Generalized-linear and logistic bandits share the likelihood but ask about covariates and prediction, with the intercept refit as one coefficient among many and never singled out as state (Filippi et al.\ 2010; Dumitrascu et al.\ 2018; Dong et al.\ 2019; Faury et al.\ 2020; Oh and Iyengar 2019); contextual linear bandits likewise give the arms their own coefficient vectors (Li et al.\ 2010; Agrawal and Goyal 2013). Nonstationary bandits forget old reward information broadly---discounting, sliding windows, change detection (Garivier and Moulines 2011; Gupta et al.\ 2011; Raj and Kalyani 2017; Burtini et al.\ 2015)---whereas OR-TS forgets one named coordinate and keeps the rest; Section~4.1 measures the difference in both regimes. Batched bandits study what the batch structure costs in regret (Perchet et al.\ 2016; Kalkanli and \"Ozg\"ur 2021); here the batch is instead the unit that makes the level identifiable. Off-policy evaluation asks a third question, the value of a fixed target policy from a logged one, and answers it with importance weights on logged propensities (Saito et al.\ 2020); OR-TS specifies a state, and the offline evaluation of an adaptive policy like it is not an importance-weighted average but a replay or a calibrated simulation, as in Section~4.3.

\runin{Same update, other fields.} Recognition in Figure~\ref{fig:agent} is the measurement update of a Kalman-type filter (Kalman 1960; Welch and Bishop 2006) whose time update discards the level and, under decay, inflates the contrast covariance; Supplement~G relates $\lambda$ to an assumed scalar process variance, without estimating that variance here. The same precision-weighted updating, with a volatility term setting the learning rate, is the Bayesian-brain account of perception (Knill and Pouget 2004; Behrens et al.\ 2007; Mathys et al.\ 2014), where $\lambda$ plays the role of volatility. Bayesian structural time series, as used to estimate the causal impact of an intervention on a single series (Brodersen et al.\ 2015), and decomposable forecasters (Taylor and Letham 2018) model the level's trajectory---local level, trend, seasonality, change points---because they must predict it, to build a counterfactual or a forecast. Basic OR-TS does not require a level forecast for allocation: concurrent arm observations identify each batch's level. Section~4.3 separately uses the fitted intercept as a forecast diagnostic. In state-space terms it is the local-level model with the level's innovation variance sent to infinity and the contrasts as the persistent component. What it needs is that concurrent exposure, not a dedicated control arm; the reference arm is a coordinate choice (Section~2.2). Where a single series is measured with no concurrent comparison, those tools apply and OR-TS does not.

\runin{Same decomposition, other field.} The level--contrast split is the bandit case of a decomposition that recurs elsewhere. In reinforcement learning an action value splits into a state value and an advantage, only the advantage orders the actions, and a baseline subtracted in a policy-gradient method leaves the expected gradient unchanged (Sutton and Barto 2018; Wang et al.\ 2016). The word \emph{state} differs between the two settings: there it is the environment's, with action-dependent transitions, whereas OR-TS's state is the belief the algorithm stores. The bandit is the case in which the baseline is never needed again: nothing the agent does changes the arms' rates, so the level can be discarded at the boundary. Once actions move the environment from one state to the next, the state value is needed to assign credit across time and is no longer a nuisance; the memory rule then applies within a state, not across the transition. Extending the agent in that direction is not attempted here.

\end{document}